\documentclass{article} 
\usepackage{iclr2021_conference,times}

\usepackage{hyperref}
\usepackage{url}
\usepackage{multirow}
\usepackage{booktabs}

\title{Spatial Frequency Bias in Convolutional Generative Adversarial Networks} 

\author{Mahyar Khayatkhoei \& Ahmed Elgammal \\
Department of Computer Science\\
Rutgers University\\
New Brunswick, NJ, USA \\
\texttt{\{m.khayatkhoei,elgammal\}@cs.rutgers.edu}
}

\usepackage{graphicx}
\usepackage{comment}
\usepackage{amsmath,amssymb,amsthm} 
\usepackage{color}
\usepackage[caption=false]{subfig}
\usepackage[title]{appendix}

\newcommand{\Nc}{\mathcal{N}}

\newcommand{\Rb}{\mathbb{R}}
\newcommand{\Nb}{\mathbb{N}}

\newcommand{\conv}{\text{Conv}}
\newcommand{\up}{\text{Up}}
\newcommand{\FT}[1]{\mathcal{F}\left\{#1\right\}}
\newcommand{\Cov}[1]{\textrm{Cov}\left[#1\right]}
\newcommand{\Var}[1]{\textrm{Var}\left[#1\right]}
\newcommand{\Sinc}{\textrm{Sinc}}

\newtheorem{propos}{Proposition}
\newtheorem{theorem}{Theorem}
\newtheorem{corollary}{Corollary}[theorem]

\newcommand{\Eq}[1]{Eq.~\eqref{#1}}
\newcommand{\Fig}{Figure}

\usepackage{xspace}
\makeatletter
\DeclareRobustCommand\onedot{\futurelet\@let@token\@onedot}
\def\@onedot{\ifx\@let@token.\else.\null\fi\xspace}

\def\eg{\emph{e.g}\onedot} 
\def\ie{\emph{i.e}\onedot}

\def\etal{\emph{et al}\onedot}
\makeatletter

\iclrfinalcopy 
\begin{document}
\maketitle

\begin{abstract}
As the success of Generative Adversarial Networks (GANs) on natural images quickly propels them into various real-life applications across different domains, it becomes more and more important to clearly understand their limitations. Specifically, understanding GANs' capability across the full spectrum of spatial frequencies, \ie beyond the low-frequency dominant spectrum of natural images, is critical for assessing the reliability of GAN generated data in any detail-sensitive application (\eg denoising, filling and super-resolution in medical and satellite images). 
In this paper, we show that the ability of convolutional GANs to learn a distribution is significantly affected by the spatial frequency of the underlying carrier signal, that is, GANs have a bias against learning high spatial frequencies. Crucially, we show that this bias is not merely a result of the scarcity of high frequencies in natural images, rather, it is a systemic bias hindering the learning of high frequencies regardless of their prominence in a dataset. Furthermore, we explain why large-scale GANs' ability to generate fine details on natural images does not exclude them from the adverse effects of this bias. Finally, we propose a method for manipulating this bias with minimal computational overhead. This method can be used to explicitly direct computational resources towards any specific spatial frequency of interest in a dataset, extending the flexibility of GANs.
\end{abstract}

\section{Introduction}
The information in an image is carried by a set of spatial frequencies, that is, a set of planar sinusoids with unique frequencies and directions. Intuitively, we associate the high frequencies with the details of an image, and the low frequencies with its general form; however, neither frequencies should be treated as more important by a generative model seeking to learn a distribution. To make this more clear, consider a two-dimensional planar cosine wave defined over a $128\times 128$ image, and assume that we sample the magnitude of this static wave from a Gaussian distribution. Whether this wave completes 64 periods across the image (\ie high spatial frequency), or 3 periods (\ie low spatial frequency), should not affect a generative model's learning of the underlying Gaussian distribution.

Convolutional Generative Advarsarial Networks (GANs)~\citep{radford2015unsupervised, goodfellow2014generative} are the foremost generative models for generating natural image distributions, and while many of their limitations have been studied from the perspective of probability theory and manifold learning~\citep{arjovsky2017towards, arora2017generalization,  khayatkhoei2018disconnected}, their spectral limitations remain mostly unexplored. The theory of GANs~\citep{goodfellow2014generative}, and its many variants~\citep{arjovsky2017wasserstein, mroueh2017fisher, nowozin2016f}, do not reveal any spectral limitation. Moreover, the recent success of large-scale GANs in learning fine details in high resolution images seems to support this notion that GANs are not sensitive to the spectral composition~\citep{karras2020analyzing, karras2018progressive}. In contrast, the progression of GAN research over the recent years reflects a constant effort for generating better \textit{details} while generating \textit{general form and color} seems to be quite easy. Can this apparent difficulty of generating details in practice be linked to an explicit spectral bias? In this work, we first mathematically reveal a spectral bias in the structure of generative convolutional neural networks (CNNs), and then empirically show its manifestation into the spatial frequency bias in GANs. Our findings can help determine which datasets are particularly prone to sub-optimal learning in a GAN training, and perhaps more importantly, which parts of a signal's information are more prone to loss. We summarize our contributions below:
\begin{itemize}
    \item We present a theorem which quantifies the linear dependency contained in the spectrum of filters used in a generative CNN, and implies a spatial frequency bias (Section~\ref{sec:theorem}).
    \item We show how GANs do not learn high frequencies as well as low frequencies in natural images, through the lens of a feature-based metric with spectral resolution (Section~\ref{sec:fid_levels}).
    \item We show the stark sensitivity of GANs' performance to the spatial frequency of the signal that carries a distribution, providing further evidence for the bias  (Section~\ref{sec:high_freq_dataset}).
    \item We explain why large-scale GANs' ability to generate very fine details on natural images does not exclude them from the adverse effects of the spatial frequency bias (Section~\ref{sec:high_res_escape}).
    \item Finally, we propose an approach for manipulating the spatial frequency bias and show its effectiveness in enhancing GANs' performance on high spatial frequencies (Section~\ref{sec:freq_shift_model}).
\end{itemize}

\begin{figure*}[t]
    \centering
    \includegraphics[trim=0 135 0 145, clip, width=0.95\textwidth]{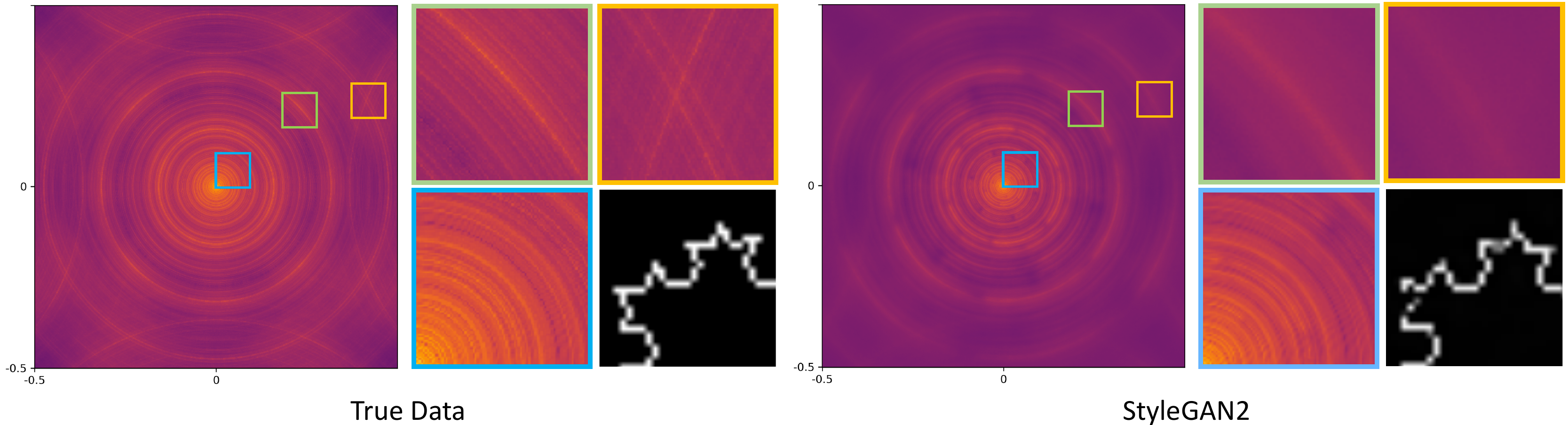}
    \includegraphics[trim=450 30 450 80, clip, width=0.037\textwidth]{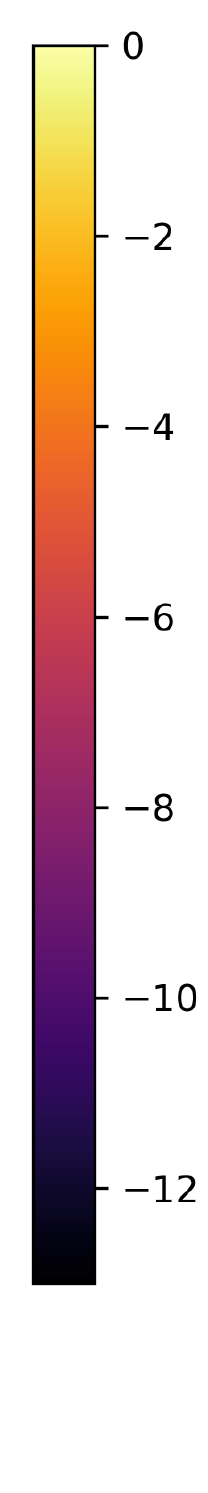}
    \caption{Average power spectrum of a large-scale GAN trained on a fractal-based dataset clearly reveals how low frequencies (closer to center) are matched much more accurately than the high frequencies (closer to corners). (Left) Average power spectrum of randomly rotated Koch snowflakes of level 5 and size $1024\times1024$. (Right) Average power spectrum of StyleGAN2 trained on the latter. A representative patch from the perimeter of true and generated fractals are also displayed.
    }
    \label{fig:koch}
\end{figure*}
\section{Spatial Frequency Components}
\label{sec:freq_comp}
According to Inverse Discrete Fourier Transform~\citep{gonzalez2002digital}, every periodic discrete 2D signal $I(x,y)$ with $x\in\{0, 1, 2, \dots, m-1\}$ and $y\in \{0, 1, 2, \dots n-1\}$, can be written as a sum of several complex sinusoids as follows:
\begin{align}
\label{eq:dft}
\begin{split}
    I(x,y) = \frac{1}{mn} \sum_{u=0}^{m-1}\sum_{v=0}^{n-1} C(u,v) e^{j2\pi(\frac{ux}{m}+\frac{vy}{n})} = \frac{1}{mn} \sum_{u=0}^{m-1}\sum_{v=0}^{n-1} C(u,v) e^{j2\pi(\hat{u}, \hat{v}).(x,y)}
\end{split}
\end{align}
We denote each complex sinusoid a \textit{spatial frequency component} which can be expressed by a vector $(u, v)$ over the pixel locations $(x, y)$. In the above equation, $C(u,v)$ is the complex amplitude of each frequency component, $(\hat{u}, \hat{v}) = (\frac{u}{m}, \frac{v}{n})$ defines the direction of propagation on the 2D plane and its magnitude defines the spatial frequency in that direction, and $m, n \in \Nb$ are the periods of $I$ in $x$ and $y$ direction respectively. Every channel of a digital 2D image can be assumed periodic beyond the image boundaries, and therefore represented by~\Eq{eq:dft}, with periods $m$ and $n$ being the length and width of the image respectively. In that case, the vector $(\hat{u}, \hat{v})$ would define the spatial frequency of a sinusoid in terms of \textit{cycles per pixel}, in x and y direction respectively, with $\hat{u},\hat{v}\in[0, 1)$. The maximum frequency in each direction is $0.5$ corresponding to the Nyquist frequency (the shortest period needs at least two pixels to be represented, hence the maximum frequency is half cycle per pixel). In favor of clarity, and without loss of generality, we will assume $\hat{u},\hat{v}\in[-0.5, 0.5)$ throughout this paper. Additionally, we loosely refer to the spatial frequency components with $|\hat{u}|$ or $|\hat{v}|$ close to $0.5$ as high frequencies, and with $\hat{u}$ or $\hat{v}$ close to $0$ as low frequencies.
Whenever displaying power spectrums $|C(\hat{u}, \hat{v})|^2$, for better visualization, we drop the dc power, apply Hann window, normalize by the maximum power, and apply $\log$, such that the most powerful frequency always has value $0$. Also, $\hat{u}$ and $\hat{v}$ are placed on horizontal and vertical axes respectively, such that low frequencies are placed close to the center, while high frequencies close to the corners.

\section{The Spatial Frequency Bias}
\label{sec:theorem}
We start by making an observation regarding the structure of a generative convolutional neural network (CNN) with regard to generating spatial frequencies. A 2D generative CNN $G(x,y;W, H^1)$, with parameters $W\in \mathcal{W}$, input features $H^1 \in \Rb^{d_0 \times d_0}$, and output space $\Rb^{d \times d}$ can be modeled as a series of convolution layers $\conv_i^l: \Rb^{d_{l-1} \times d_{l-1}} \rightarrow \Rb^{d_l \times d_l}$ as follows~\footnote{Transposed Conv layers are sufficiently represented by an appropriate choice of the $\up(.)$ operator.}:
\begin{align}
\label{eq:cnn}
    H^{l+1}_i = \conv_i^l(H^l) = \sum_c F^l_{ic} * \up \left( \sigma(H^l_c) \right)
\end{align}
where $l$ indices the layer (depth), $i$ the output channels, $c$ the input channels, $F^l_{ic} \in \Rb^{k_l \times k_l}$ is a parametric 2D filter, $\up(.)$ denotes the upsampling operator, and $\sigma(.)$ is a non-linearity.
If we restrict $\sigma$ to rectified linear units (ReLU), then in a neighborhood of almost any parameter $W$, we can consider the combined effect of $\up(\sigma(.))$ as a fixed linear operation:

\begin{propos}
At any latent input $H^1$ of a finite size ReLU-CNN, almost everywhere on the parameter space, there exists a neighborhood in which ReLUs are equivalent to fixed binary masks~\footnote{\label{fnote}Proof in Appendix.}. 
\end{propos}

Therefore, in this neighborhood, improving the output spectrum is only achievable through adjusting the spectrum of filters $F^l_{ic}$. Intuitively, the filters try to \textit{carve out} the desired spectrum out of the input spectrum which is distorted by ReLUs (as binary masks), and aliased by upsampling. In the following theorem, we investigate how freely can these filters adjust their spectrum. Specifically, we will show how the filter size $k_l$ and the spatial dimension $d_l$ of a Conv layer affect the correlation in the spectrum of its filters. Note that more correlation in a filter's spectrum means more linear dependency, and thus reduces its effective capacity, in other words, the filter can not freely adjust specific frequencies without affecting the adjacent correlated frequencies.

\begin{theorem}
Let $U=\FT{F^l_{ic}}(u_0, v_0)$ and $V=\FT{F^l_{ic}}(u_1, v_1)$ be any two spatial frequency components on the spectrum of any 2D filter of the $l$-th Conv layer with spatial dimension $d_l$ and filter size $k_l$, at any point during training. Assuming i.i.d. weight initialization, the magnitude of complex correlation coefficient between $U$ and $V$ is given by~\footnotemark[\value{footnote}]:

\begin{align}
    \label{eq:cnn_corr}
    \begin{split}
    &|\textrm{corr}(U, V)| = \left|\frac{\Sinc(u_0-u_1, v_0-v_1)}{k_l^2} \right|\\
    &\textrm{s.t.} \quad \Sinc(u,v) = \frac{\sin(\frac{\pi uk_l}{d_l})\sin(\frac{\pi vk_l}{d_l})}{\sin(\frac{\pi u}{d_l})\sin(\frac{\pi v}{d_l})} \quad d_l, k_l \in \Nb \quad \textrm{and} \quad 1 < k_l \leq d_l\\
    \end{split}
\end{align}
\end{theorem}

\begin{corollary}
If $U$ and $V$ are two diagonally adjacent spatial frequency components of $F^l_{ic}$, then:
\begin{align}
\label{eq:cnn_corr_cor1}
    |\textrm{corr}(U, V)| = \frac{\sin^2(\frac{\pi k_l}{d_l})}{k_l^2\sin^2(\frac{\pi}{d_l})}
\end{align}
\end{corollary}

\begin{figure*}[t]
    \centering
    \includegraphics[trim=30 10 50 50, clip, width=0.45\textwidth]{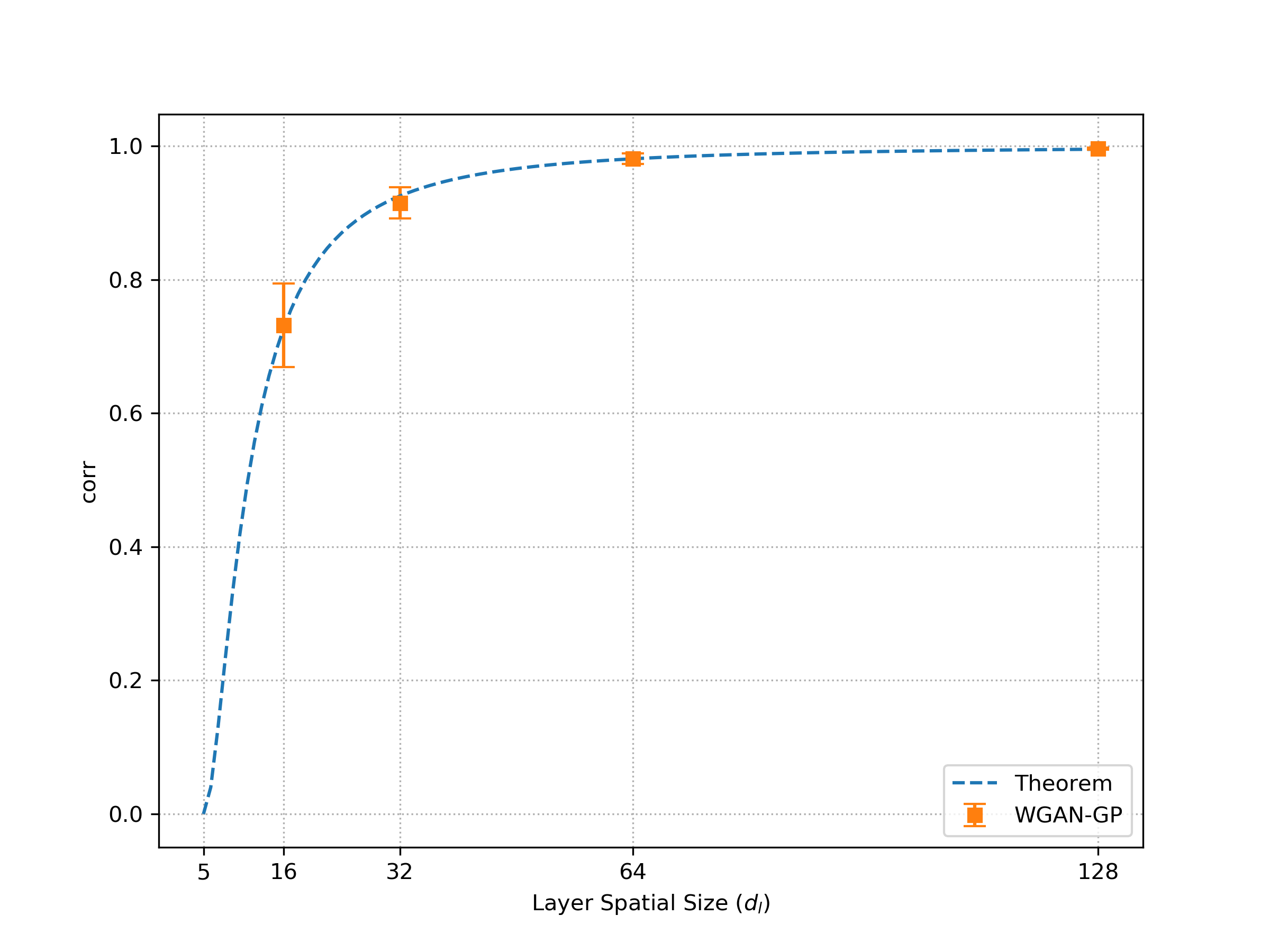}
    \includegraphics[trim=180 40 160 80, clip, width=0.52\textwidth]{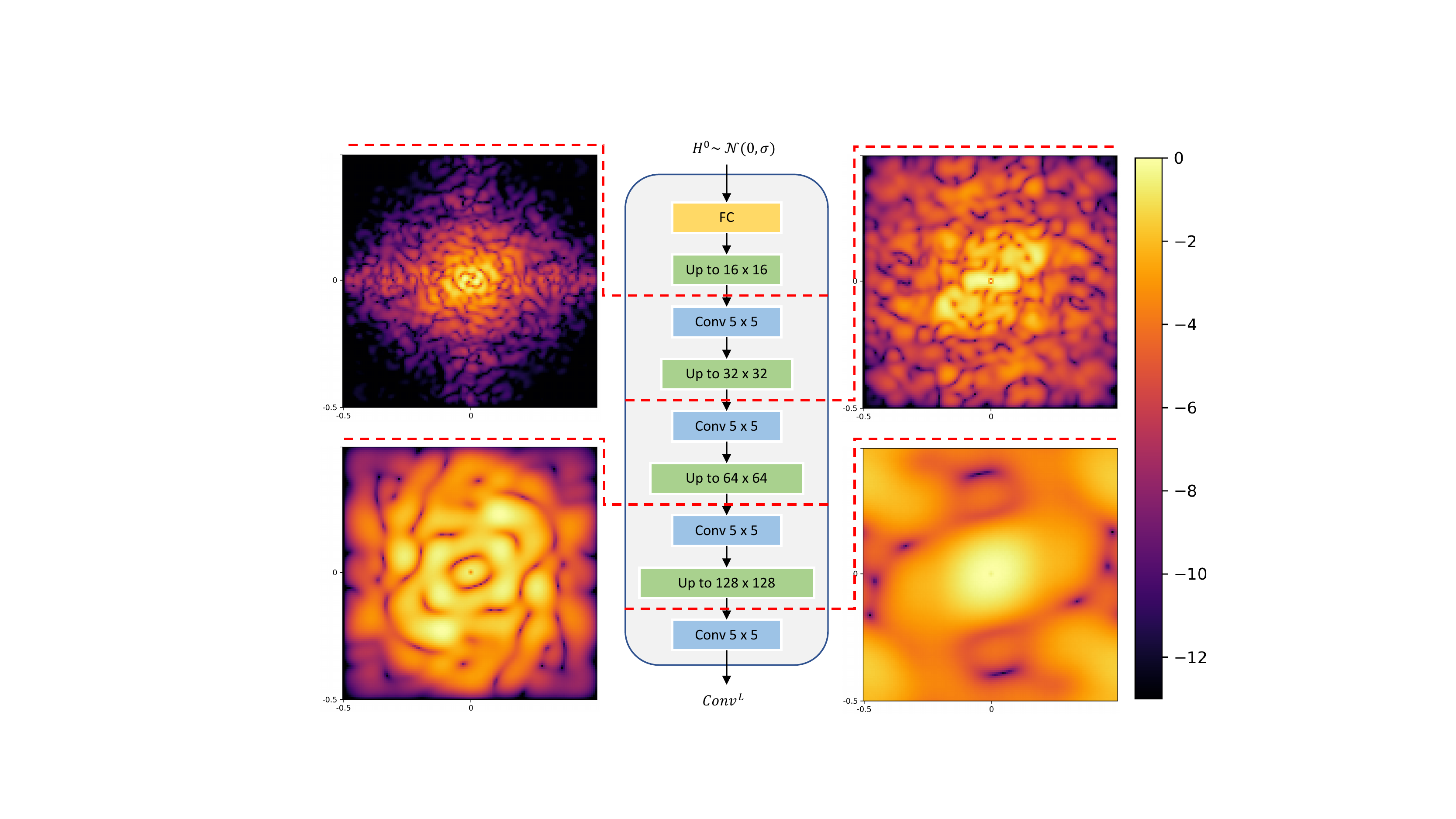}
    \caption{(Left) Dashed blue line shows the predicted correlation between diagonally adjacent spatial frequencies in \Eq{eq:cnn_corr_cor1} for different layers ($d_l$) given a fixed filter size ($k_l=5$), and the orange points show its empirical evaluation (average correlation with one standard deviation computed over the filters of WGAN-GP trained on CelebA). (Right) The impulse response spectrum of effective filters operating on each layer in WGAN-GP trained on CelebA. Notice how the spectrum of the effective filters that operate on inner layers, \ie control the generation of low frequencies (top row), are much sharper than the ones that operate on outer layers, \ie control the generation of high frequencies (bottom row). Smoothness is an indication of larger correlation in the effective filters.
    }
    \label{fig:theorem}
\end{figure*}

Now, in each Conv layer, note that the maximum spatial frequency that can be generated is limited by the Nyquist frequency, that is, $\conv^l$ can only adjust image spatial frequencies in $[0, \frac{d_l}{2d}]$ without aliasing\footnote{Aliasing here refers to the process of generating high frequencies by replicating low frequencies in the expanded spectrum introduced through upsampling. Since this makes duplicates in several high frequency bands of the spectrum, it's ability to control high frequencies is minimal. Many up-sampling approaches, such as bi-linear or nearest-neighbors, explicitly attenuate such aliased frequencies.}. This means that high frequencies are primarily generated by the CNN's outer layers which have larger spatial dimension $d_l$. According to \Eq{eq:cnn_corr_cor1}, given a fixed filter size $k_l$, the larger the $d_l$, the larger the correlation in the filter's spectrum (see the equation's graph in \Fig{}~\ref{fig:theorem}), and consequently the smaller its effective capacity. Therefore, the outer layers responsible for generating high frequencies are more restricted in their spectrum compared to the inner layers with smaller $d_l$. Additionally, note that while only outer layers can generate high frequencies without aliasing, the low frequencies can be generated by all layers without aliasing. Therefore, the spatial size of the effective filter operating on low frequencies will always be larger than that operating on high frequencies. That means, even if larger filter size $k_l$ is used in the outer layers to counteract the effect of larger $d_l$, the low frequencies will still be enjoying a larger end-to-end filter size compared to high frequencies, hence less correlation (see the spectrums of effective filters in \Fig{}~\ref{fig:theorem}).

In light of these theoretical observations, we hypothesize that convolutional GANs contain a systemic bias against generating high frequencies. We particularly choose to study this bias in GANs for two reasons: first, the spectral limitations of GANs are poorly understood and it is critical to investigate such limitations as their application rapidly expands across many tasks; second, unlike L2-reconstruction and Variational Autoencoders, GANs do not contain any spectral bias inherent to their objective functions, and can therefore provide a nice framework for observing a spectral bias in generative CNNs. In the remainder of this section we will empirically investigate our hypothesis on three popular convolutional GANs: WGAN-GP~\citep{gulrajani2017improved} serves as a simple but fundamental GAN model; and Progressively Growing GAN (PG-GAN)~\citep{karras2018progressive} and StyleGAN2~\citep{karras2020analyzing} serve as state-of-the-art models with large capacity and complex structure, incorporating state-of-the-art normalization and regularization techniques. In order to have an approximately consistent CNN structure, in all models, the generative CNNs start from a 512 dimensional latent space, and after the $8\times8$ spatial dimensions, the channel size is consecutively halved every time the spatial dimensions are doubled (note that this makes the total capacity of PG-GAN and StyleGAN2 slightly smaller than original, but this does not affect an empirical investigation of a spectral bias in convolutional GANs, since we are not comparing these models with each other, nor trying to beat state-of-the-art performance). See Appendix for more details.

\subsection{FID Levels}
{
\setlength{\tabcolsep}{2pt}
\renewcommand{\arraystretch}{1}
\begin{figure*}[t!]
    \centering
    \begin{tabular}{cccc}
        & WGAN-GP & PG-GAN & StyleGAN2\\
        \raisebox{4\normalbaselineskip}[0pt][0pt]{\rotatebox[origin=c]{90}{CelebA}} & \includegraphics[trim=30 10 50 50, clip, width=0.3\textwidth]{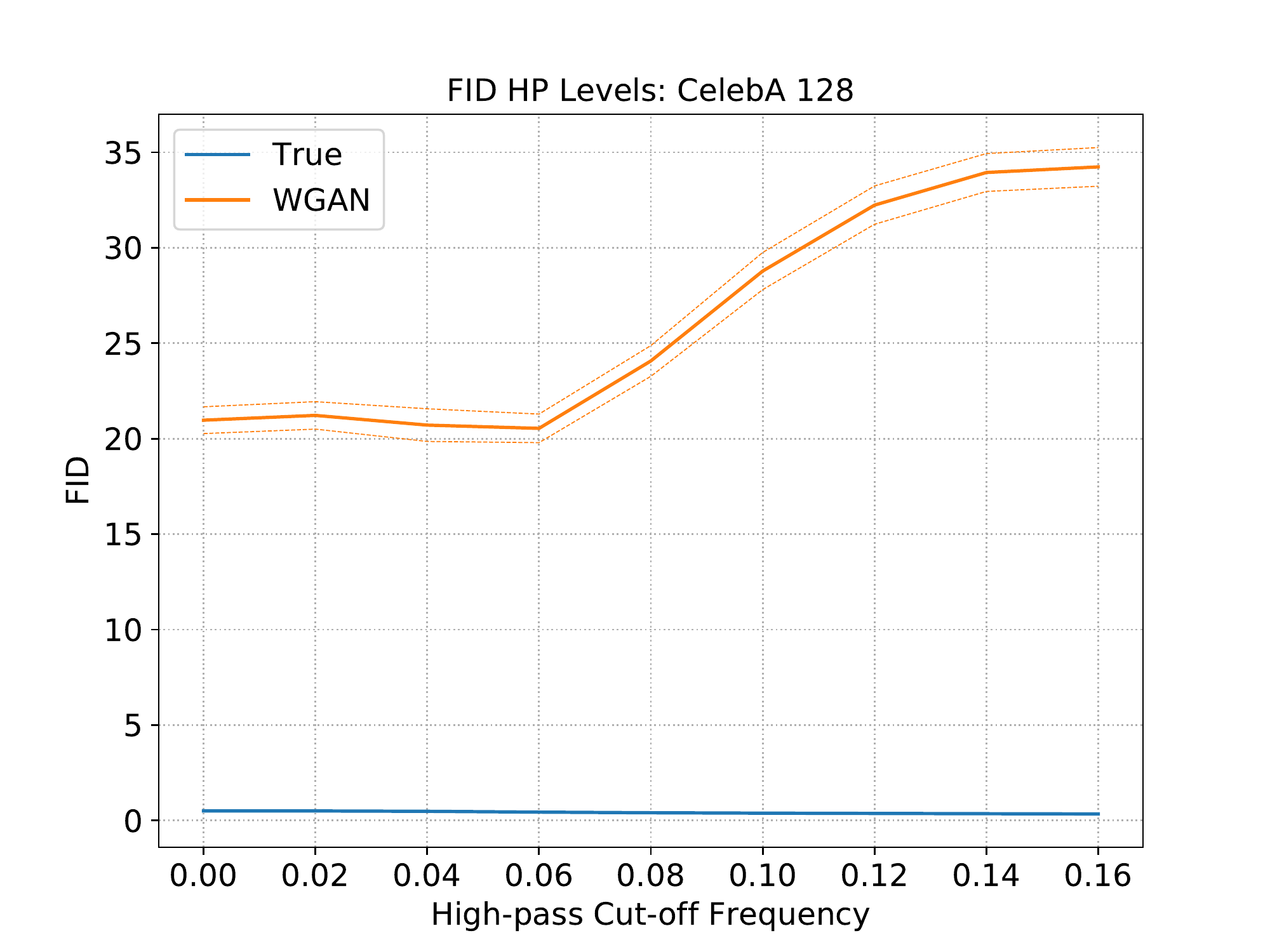} & \includegraphics[trim=30 10 50 50, clip, width=0.3\textwidth]{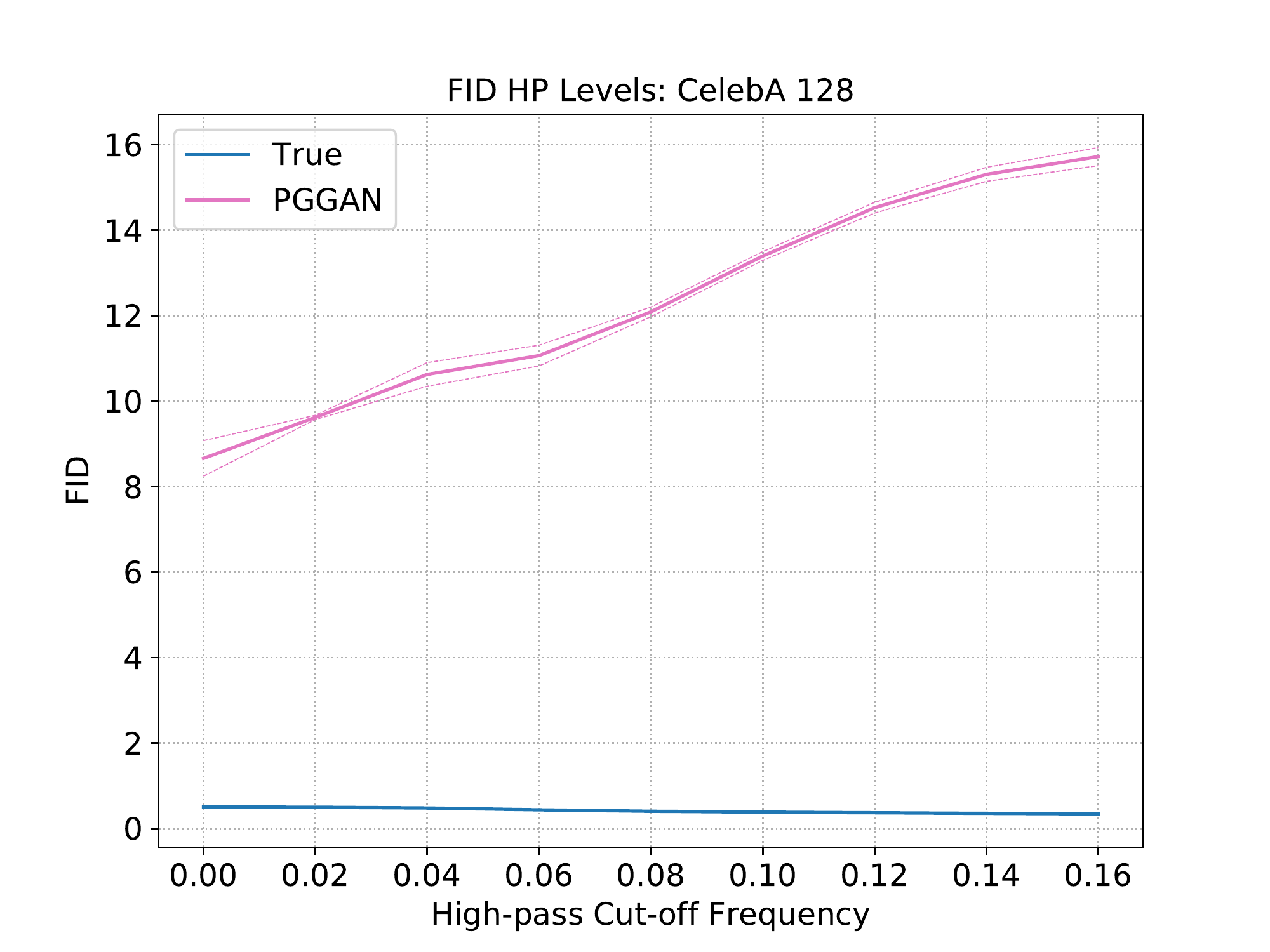} &  \includegraphics[trim=30 10 50 50, clip, width=0.3\textwidth]{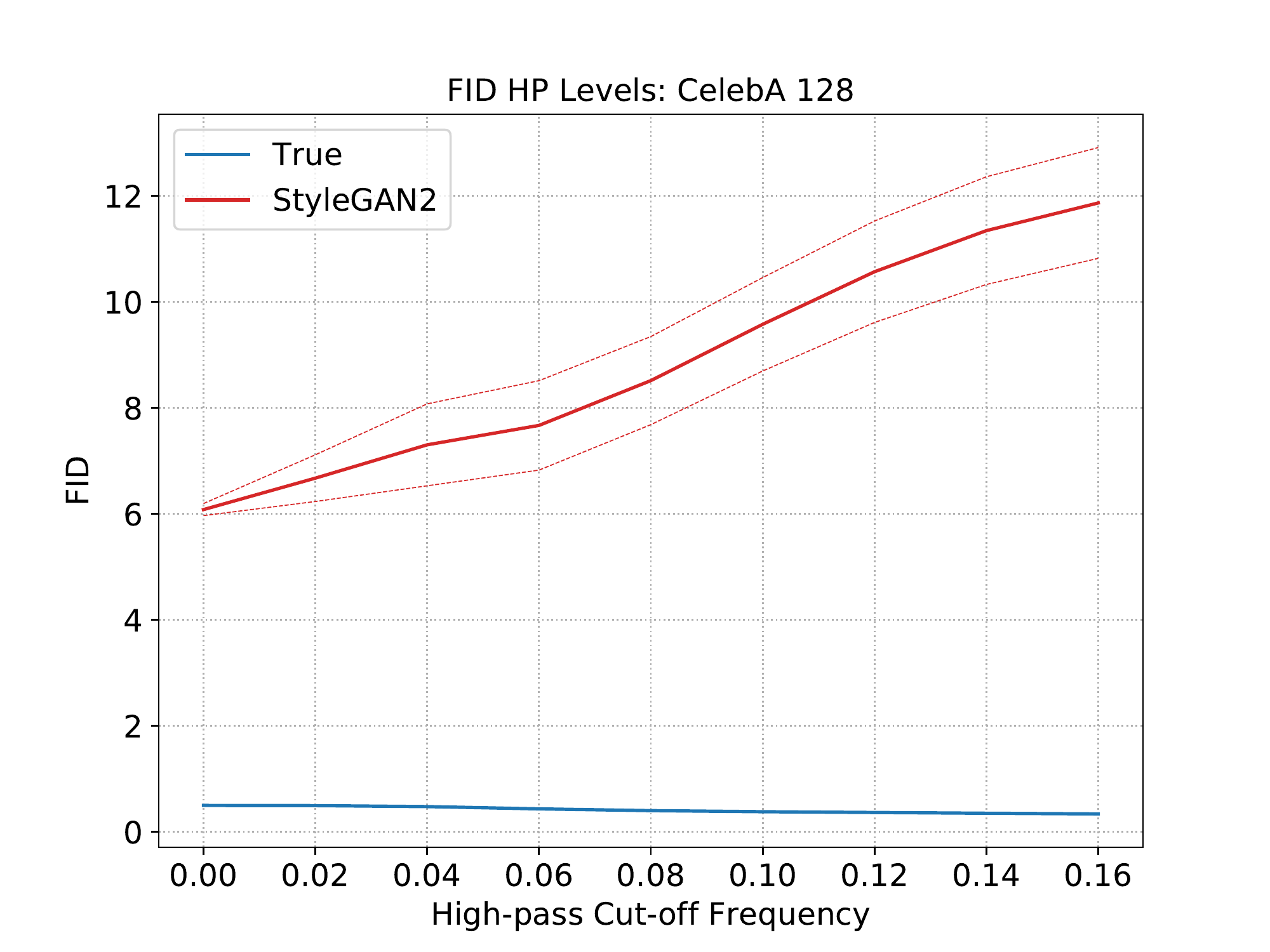}\\
        \raisebox{4\normalbaselineskip}[0pt][0pt]{\rotatebox[origin=c]{90}{Bedrooms}} & \includegraphics[trim=30 10 50 50, clip, width=0.3\textwidth]{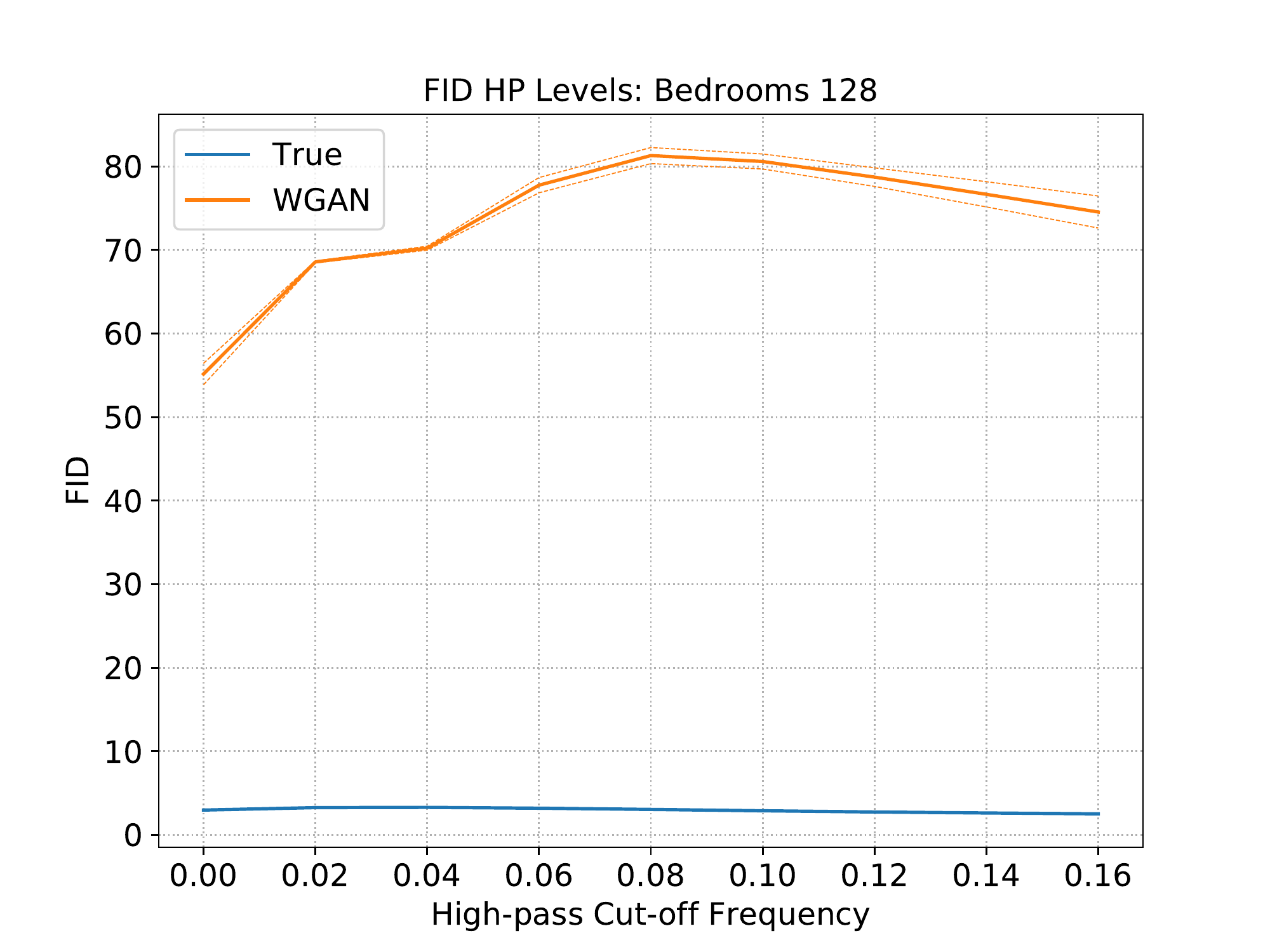} & \includegraphics[trim=30 10 50 50, clip, width=0.3\textwidth]{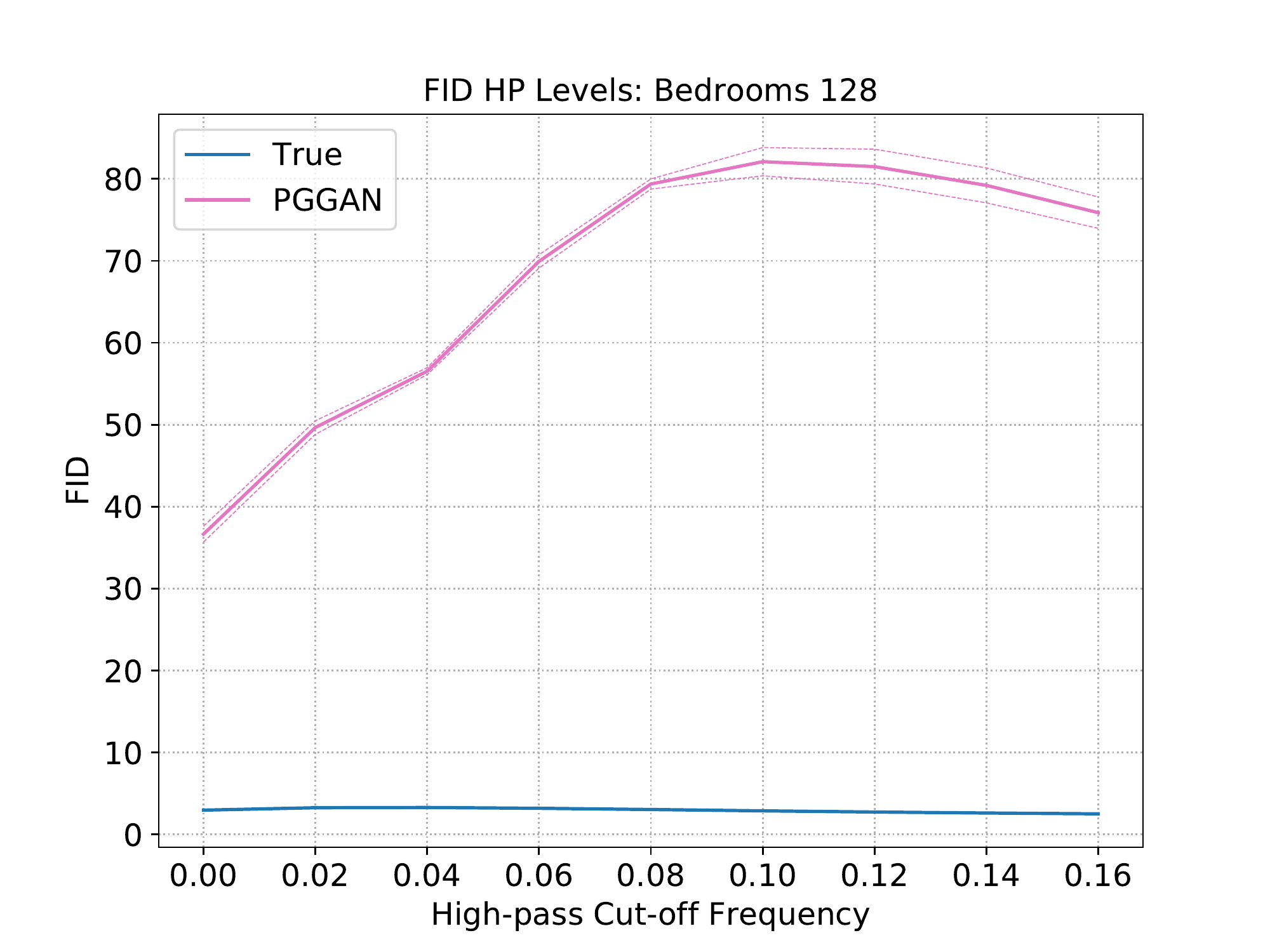} & \includegraphics[trim=30 10 50 50, clip, width=0.3\textwidth]{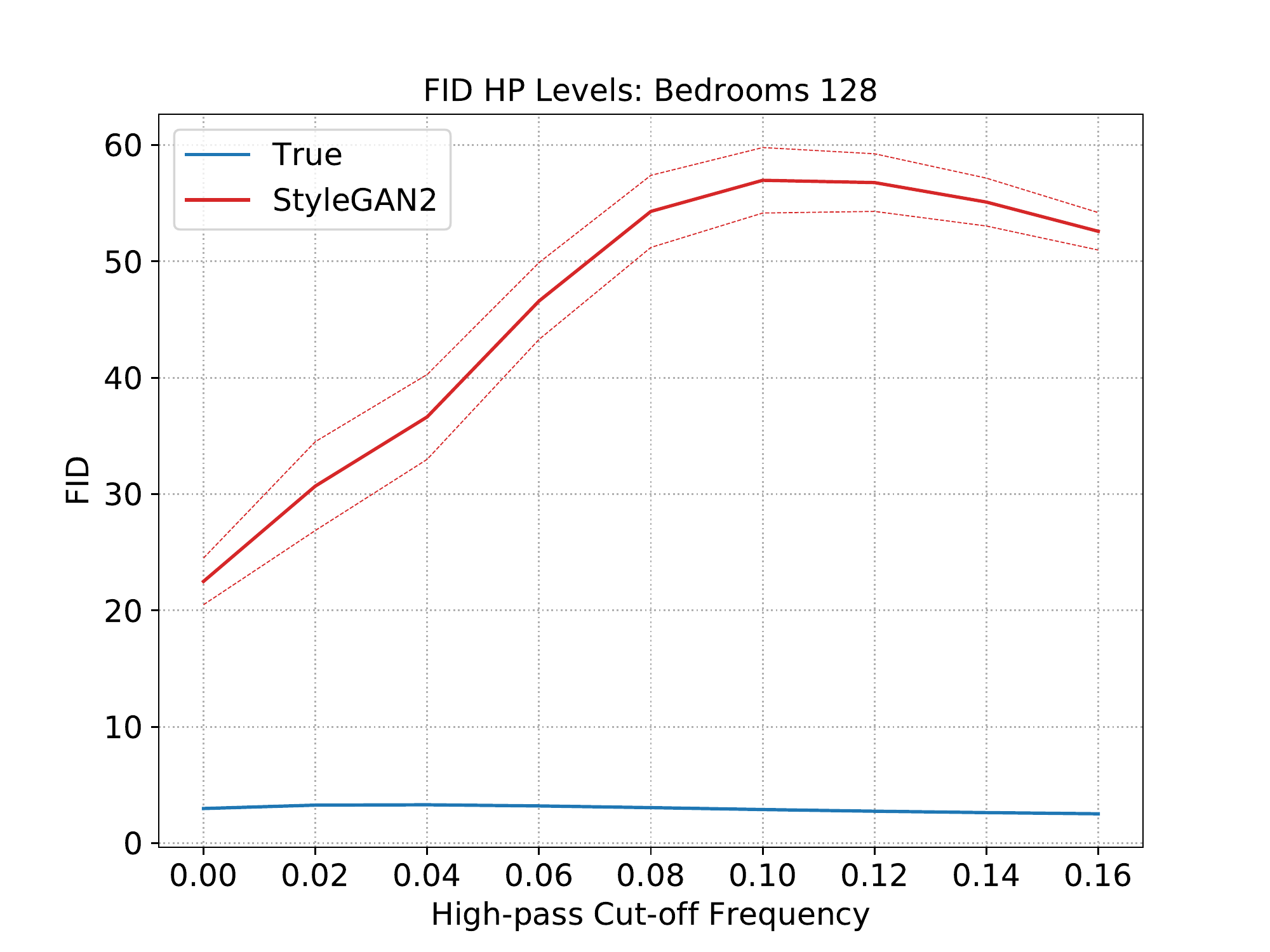}
    \end{tabular}
    \caption{FID Levels of GANs trained on CelebA and LSUN-Bedrooms. The farther to the right on the horizontal axis, the more low frequencies are removed prior to FID computation. Notice the transient increase in FID (worsening performance) as low frequencies are removed. All plots will eventually decline to zero if we continue removing frequencies. In all figures, the \textit{blue curve} depicts the True FID Levels of the corresponding dataset as a baseline. All figures show average FID with one standard deviation error bars (\textit{dashed line}), over three random training runs.
    }
    \label{fig:freq_bias}
\end{figure*}
}
\label{sec:fid_levels}
Direct comparison between the average power spectrums of generated and true images is not a very informative measure of spectral performance, since a numerical difference in average power spectrums does not necessarily mean a meaningful difference in discernible image features. Instead, we propose an extended version of Frechet Inception Distance (FID)~\citep{heusel2017gans}, denoted \textit{FID Levels}, in which we plot FID between two sets of images after gradually removing low frequency bands from both sets. The intuition is to observe the change in the GAN's performance as frequency bands are gradually removed.
FID measures the differences between the features extracted from two image distributions: the larger the FID, the larger the mismatch between the two distributions. We chose FID because of its sensitivity to a wide range of high and low frequency features~\citep{borji2019pros, lucic2018gans}. Each point on the FID Levels plot shows FID computed after applying a high-pass Gaussian filter, with the cut-off specified on the horizontal axis, to both the generated and the true images. As a baseline for comparison, we also compute FID Levels between two disjoint subsets of the true images, denoted \textit{True FID Levels}, demonstrating the ideal GANs' performance. For completeness, as a direct measure of differences between power spectrums, we also report \textit{Leakage Ratio (LR)} in our experiments, which computes total variation~\citep{gibbs2002choosing} between the generated and true average power spectrums normalized into density functions. Intuitively, LR shows what percentage of the power has leaked away between two power spectrums.

Removing frequency bands removes information from both image sets, and at the limit, where all the spatial frequencies are removed, the two image sets would look exactly the same, therefore, we expect a declining plot in FID Levels overall. However, if the GAN has matched certain frequency bands better than the others, removing those bands from images would cause an increase in FID. \Fig{}~\ref{fig:freq_bias} shows FID Levels of GANs trained on two $128\times128$ image datasets: CelebA~\citep{liu2015celeba} and LSUN-Bedrooms~\citep{yu15lsun}. The GANs exhibit an increase in FID Levels on both datasets, which shows that removing low frequencies has emphasized the mismatch between the true and the generated images, hence a bias against learning high frequencies. Without such a bias, we would only observe a decline in FID as low frequencies are gradually removed. However, since much of the information in natural images is concentrated at low frequencies~\citep{field1987relations}, this bias might be attributed to the scarcity of high frequencies during training. We will show that this cannot be the explanation in the following subsection.

\subsection{High Frequency Datasets}
\label{sec:high_freq_dataset}
If GANs are not biased against high frequencies, their performance should remain indifferent to shifting the frequency contents of the datasets. In other words, whether the information in a dataset is primarily carried by high frequencies or low frequencies should not affect how well GANs can learn the underlying image distribution. In order to test this hypothesis, we can create high frequency versions of CelebA and LSUN-Bedrooms by applying a frequency shift operator, that is, multiplying every image in each dataset with $\cos(\pi (x+y))$ channel-wise, to create Shifted CelebA (ScelebA) and Shifted LSUN-Bedrooms (SBedrooms) respectively. In effect, all we are doing is swapping the low and high frequency contents of these datasets. Note that the frequency shift operator is a homeomorphism and therefore the distributions of SCelebA and SBedrooms have the same topological properties as CelebA's and LSUN-Bedroom's, and therefore the GANs' performance should remain unchanged from a purely probabilistic perspective.

Table~\ref{tab:high_freq_dataset} compares the GANs' performance on SCelebA and SBedrooms versus the original CelebA and LSUN-Bedrooms.\footnote{In SCelebA and SBedrooms, true and generated images are re-shifted before computing FID so that the values are comparable with the FID results on CelebA and LSUN-Bedrooms} Their performance has worsened significantly (larger FID and LR) on the high frequency datasets, showing that the GANs perform considerably better when the same image distribution is carried primarily by low frequencies. This observation rejects our earlier hypothesis, that is, the GANs' performance is not indifferent to frequency shift. Additionally, this shows that the bias against high frequencies we observed in Section~\ref{sec:fid_levels} cannot be explained by the scarcity of high frequencies in natural images: even though the unbalancedness in the distribution of power has remained unchanged in the high frequency versions of the datasets, the GANs' performance has worsened significantly. We conclude that this bias is indeed a ``spatial frequency bias'' against high frequencies, regardless of how abundant or scarce they are in the dataset.
{
\setlength{\tabcolsep}{7pt}
\begin{table*}
\begin{center}
\caption{Performance drop (increase in both FID and LR) when GANs are trained on the high frequency versions of CelebA and LSUN-Bedrooms, denoted SCelebA and SBedrooms respectively. Average measures with standard deviation (sd) are reported over three random training runs.}
\vspace{\baselineskip}
\label{tab:high_freq_dataset}
\begin{tabular}{ll|ll|ll}
\toprule\noalign{\smallskip}
\multicolumn{1}{c}{Model} & \multicolumn{1}{c}{Measure} & \multicolumn{1}{c}{CelebA} & \multicolumn{1}{c}{SCelebA} & \multicolumn{1}{c}{Bedrooms} & \multicolumn{1}{c}{SBedrooms}\\
\midrule
\multirow{2}{*}{WGAN-GP} & FID & $20.97$~sd~$0.70$ & $328.72$~sd~$9.70$ & $55.14$~sd~$1.29$ & $283.02$~sd~$7.06$\\
& LR (\%) & $2.29$~sd~$0.31$ & $59.04$~sd~$5.09$ & $1.99$~sd~$0.48$ & $42.42$~sd~$4.32$\\
\midrule
\multirow{2}{*}{PG-GAN} & FID & $8.66$~sd~$0.41$ & $23.12$~sd~$2.08$ & $36.65$~sd~$0.97$ & $69.03$~sd~$10.28$\\
& LR (\%) & $1.06$~sd~$0.21$ & $3.93$~sd~$0.70$ & $1.51$~sd~$0.29$ & $3.12$~sd~$0.16$\\
\midrule
\multirow{2}{*}{StyleGAN2} & FID & $6.08$~sd~$0.11$ & $343.57$~sd~$53.59$ & $22.49$~sd~$2.00$ & $260.84$~sd~$4.03$\\
& LR (\%) & $1.55$~sd~$0.42$ & $43.03$~sd~$34.10$ & $1.28$~sd~$0.19$ & $7.32$~sd~$1.86$\\
\bottomrule
\end{tabular}
\end{center}
\end{table*}
}

\subsection{A High Resolution Escape}
\label{sec:high_res_escape}
It is key to note that whether a signal contains high spatial frequencies or not is directly related to its sampling rate. For example, consider the continuous-valued image of a bird formed on a camera's sensor, whose feathers change color from white to black 64 times over the length of the image. If this image is sampled into a $128\times128$ picture, the feathers would form a high frequency component ($\frac{1}{2}$ cycles per pixel). If we instead sample the same image into a $1024\times1024$ picture, the feathers, still changing color 64 times over the length of the image, now create pixel value changes every 16 pixels, forming a low frequency component ($\frac{1}{16}$ cycles per pixel). Therefore, one solution to the spatial frequency bias is to simply use data at a very high resolution such that no high frequency component remains. We hypothesize that this can partly explain the success of large-scale GANs at generating fine details in high resolution natural image datasets~\citep{karras2020analyzing, karras2018progressive, brock2018large}. These models, by introducing novel techniques, are able to train very large networks on high resolution images, in which details are no longer ``fine'' as far as the generator is concerned. However, this does not mean that they perform similarly on all frequencies and do not suffer from the spatial frequency bias. To illustrate this, we train the large-scale StyleGAN2 (config-e) on a $1024\times1024$ dataset of randomly rotated 5th-level Koch snowflakes, a high resolution dataset that does contain prominent low and high frequencies by design. \Fig{}~\ref{fig:koch} shows how this model clearly still exhibits the signs of a bias against high frequencies.
Thus, sampling at a high resolution is essentially escaping the spatial frequency bias, not solving the bias, and such an escape is not always practical. Training GANs on higher resolution images requires exponentially more training resources, which can quickly make the training impractical if the data inherently contains very high frequencies. Therefore, it is important to explore alternative approaches for dealing with the bias. 

\section{Frequency Shifted Generators (FSG)}
\label{sec:freq_shift_model}
In section~\ref{sec:theorem}, we observed that GANs have a spatial frequency bias, favoring the learning of low frequencies, however, \textit{is it possible to manipulate this bias such that it favors other frequencies?} If so, this would make it possible to explicitly target specific frequencies of interest in a dataset. In this section, we show how this can be achieved without any increase in training resources.
Instead of inherently generating high frequencies, a generative model $G(x,y)$ can first generate a signal with prominent low frequencies and then transform the signal such that these prominent frequencies are shifted towards a desired frequency $(\hat{u}_t, \hat{v}_t)$. This can be achieved by a frequency shift operator:
\begin{align}
\label{eq:freq_shift}
\begin{split}
    G(x,y) e^{j2\pi (\hat{u}_t x + \hat{v}_t y)} = \frac{1}{mn}\sum_{u=0}^{m-1}\sum_{v=0}^{n-1} C(u, v) e^{j2\pi(\hat{u}+\hat{u}_t, \hat{v}+\hat{v}_t).(x,y)}
\end{split}
\end{align}
where $\hat{u}_t,\hat{v}_t \in [-0.5, 0.5)$. After the frequency shift, the frequency components previously close to $(0,0)$ are now placed close to $(\hat{u}_t, \hat{v}_t)$. However, since $G$ is generating a real signal and the spectrum of real signals are symmetric, it can not sufficiently represent a high frequency band, that is, $G$ can only represent symmetric frequency bands. Note that while natural images are real signals and have symmetric spectrum with respect to zero, a specific band of their spectrum is not necessarily symmetric. In order to generate a non-symmetric frequency band, we can use two neural networks to generate a real image ($G_r$) and an imaginary image ($G_i$), which together compose the complex generated image ($G_c$). The complex image is then shifted to $(\hat{u}_t, \hat{v}_t)$ according to \Eq{eq:freq_shift} to construct the shifted generator $G_s$ as follows:
\begin{align}
\begin{split}
    &G_s(x,y) = G_c(x,y) e^{j2\pi (\hat{u}_t x + \hat{v}_t y)} = \left[ G_r(x,y)+jG_i(x,y) \right] e^{j2\pi (\hat{u}_t x + \hat{v}_t y)}
\end{split}
\end{align}

The real part of $G_s$ is now generating an image which can sufficiently represent any frequency band, and has a spatial frequency bias favoring the desired component $(\hat{u}_t, \hat{v}_t)$:
\begin{align}
\label{eq:shift_gen}
\begin{split}
    \Re\left[G_s(x,y)\right] =
    G_r(x,y) \cos(2\pi (\hat{u}_t x+\hat{v}_t y))-G_i(x,y) \sin(2\pi (\hat{u}_t x+\hat{v}_t y))
\end{split}
\end{align}

\begin{figure*}[t]
    \centering
    \includegraphics[trim=0 80 0 120, clip, width=1\textwidth]{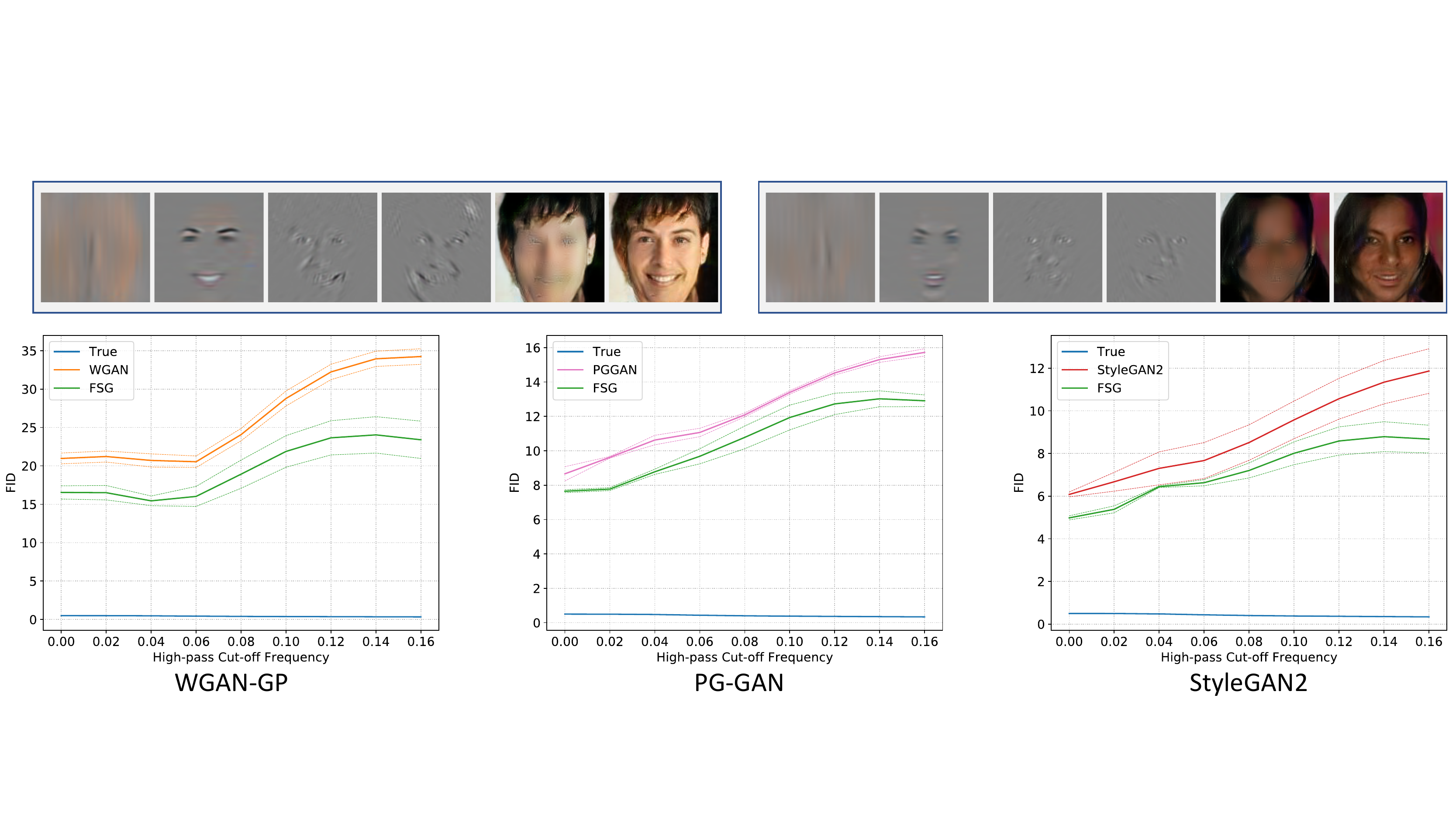}
    \caption{(Top) Two samples from WGAN-GP when enhanced by adding multiple FSGs. In each sample, from left to right, the outputs correspond to the FSG with $(\hat{u}_t, \hat{v}_t)$ at $(\frac{1}{16}, 0)$, $(0, \frac{1}{16})$, $(-\frac{1}{16}, \frac{1}{16})$, and $(\frac{1}{16}, \frac{1}{16})$, the WGAN-GP's main generator, and the final compound output (sum of all the preceding generators). Notice how each FSG has learned to focus on specific spatial frequencies, without any explicit supervision during training. See Appendix for more samples. (Bottom) Improvement in the FID Levels of GANs when enhanced by multiple FSGs, trained on CelebA.
    }
    \label{fig:fsg_multi}
\end{figure*}

Frequency Shifted Generators (FSGs) can be used to efficiently target specific spatial frequency components in a dataset. Table~\ref{tab:fsg} shows the results of training GANs using FSG with $(\hat{u}_t, \hat{v}_t) = (\frac{1}{2}, \frac{1}{2})$ on SCelebA and SBedrooms. The use of FSG has considerably improved the GANs' performance on these high frequency datasets, without any increase in training resources. This also provides an interesting insight: the discriminator is able to effectively guide a capable generator towards learning high frequencies, therefore, the spatial frequency bias must be primarily rooted in the GAN's generator and not the discriminator. Moreover, multiple shifted generators, with smaller network capacity, can be added to the main generator of GANs to improve performance on specific frequencies (\eg see \Fig{}~\ref{fig:fsg_multi}). Interestingly, the added FSGs specialize towards their respective target frequency $(\hat{u}_t, \hat{v}_t)$, without any explicit supervision during training. This provides further evidence of the spatial frequency bias: if unbiased, the added FSGs would have no incentive to specialize towards any specific frequency.
{
\setlength{\tabcolsep}{7pt}
\renewcommand{\arraystretch}{1}
\begin{table*}[t]
\begin{center}
\caption{Performance gain (decrease in both FID and LR) on the high frequency datasets achieved by using FSG in GANs. Average measures with sd are reported over three random training runs.}
\vspace{\baselineskip}
\label{tab:fsg}
\begin{tabular}{l|ll|ll}
\toprule\noalign{\smallskip}
\multicolumn{1}{c}{} & \multicolumn{2}{c}{SCelebA} & \multicolumn{2}{c}{SBedrooms}\\
\midrule
\multicolumn{1}{c}{Model} & \multicolumn{1}{c}{FID} & \multicolumn{1}{c}{LR (\%)} & \multicolumn{1}{c}{FID} & \multicolumn{1}{c}{LR (\%)}\\
\midrule
WGAN-GP & $328.72$~sd~$9.70$ & $59.04$~sd~$5.09$ & $283.02$~sd~$7.06$ & $42.42$~sd~$4.32$\\
WGAN-FSG & $20.70$~sd~$0.44$ & $1.93$~sd~$0.57$ & $59.81$~sd~$1.64$ & $1.80$~sd~$0.28$\\
\midrule
PG-GAN & $23.12$~sd~$2.08$ & $3.93$~sd~$0.70$ & $69.03$~sd~$10.28$ & $3.12$~sd~$0.16$\\
PG-GAN-FSG & $17.91$~sd~$0.74$ & $2.96$~sd~$0.55$ & $54.64$~sd~$0.26$ & $2.67$~sd~$0.75$\\
\midrule
StyleGAN2 & $343.57$~sd~$53.59$ & $43.03$~sd~$34.10$ & $260.84$~sd~$4.03$ & $7.32$~sd~$1.86$\\
StyleGAN2-FSG & $7.17$~sd~$0.07$ & $1.41$~sd~$0.10$ & $67.85$~sd~$2.38$ & $1.82$~sd~$0.27$\\
\bottomrule
\end{tabular}
\end{center}
\end{table*}
}

\section{Related Works}
\label{sec:related_works}
\textbf{Spectral Theorems of Neural Networks.} \citet{pmlr-v97-rahaman19a} have recently studied the spectral properties of fully connected neural networks that use ReLU activations, and discovered a bias against learning high frequencies in a regression task. However, while Rahaman~\etal define a frequency component as a periodic change in a single output of the network with respect to changes in the input space, we define a frequency component as a periodic change across the adjacent outputs of the network. Note that these two notions of frequency are independent by definition, that is, one notion can be mathematically defined while the other is not and vice versa (by restricting the input or output space to a single point), therefore a bias in one can not imply a bias in the other.

\textbf{Spectral Limitations.}
Spectral limitations have been observed in different tasks when using generative CNNs. In L2-reconstruction tasks, there is a bias against low-power frequencies (which coincide with high frequencies in natural images) primarily due to the vanishing gradient of the L2 loss near zero, which is often counteracted by emphasizing the low-power frequencies through additional loss terms~\citep{deng2020learning, li2018spectral, ulyanov2018deep}. In Auto Encoders and Variational Auto Encoders (VAEs), there is a similar bias, primarily due to the distribution assumptions in their objective~\citep{huang2018introvae, larsen2016autoencoding}. In contrast, GAN's objective does not impose any such spectral limitations. In theory, GANs must be able to learn any suitable distribution regardless of the carrier spatial frequencies. Therefore, while a systemic spectral bias in generative CNNs could be obscured by the inherent spectral biases of AEs, VAEs and L2 tasks, the GANs provide a clear lens for observing such biases.
Most relevant to our work, two concurrent works~\citep{dzanic2019fourier, durall2020watch} have recently shown that the discrepancies in high frequencies can be utilized to easily distinguish GAN generated images from natural images. \citet{durall2020watch} connect the discrepancies to the aliasing introduced by the up-sampling operator in the expanded spectrum. However, note that if the up-sampling does not introduce aliased power in the expanded frequency spectrum, the following layer will not have any power to filter-out and shape the desired higher frequencies. In contrast, in Theorem 1, we show that it is the linear dependencies in the Conv filter's spectrum that causes spectral limitations, and explain why this affects high frequencies more severely than low frequencies, causing a systemic bias against high frequencies, not just artifacts in the power spectrum.

\textbf{Quantitative Metrics.}
The prevalent metrics for evaluating GANs, most notably Inception Score~\citep{salimans2016improved}, FID~\citep{heusel2017gans}, and MS-SSIM~\citep{odena2017conditional}, consider all spatial frequency components at once when comparing GAN generated and true images, thus lacking any spectral resolution. Most relevant to our proposed metric, \citet{karras2018progressive} propose computing sliced Wasserstein distance between patches extracted from true and generated images at different levels of a Laplacian pyramid (SWD). Interestingly, evaluating GANs with SWD shows approximately similar performance across frequency bands~\citep{karras2018progressive}. We conjecture that this discrepancy comes from the fact that small differences between patches in the pixel space, can result in large differences in the more meaningful feature space used by FID. As such, FID Levels is a more informative measure of GANs' performance across spatial frequencies.

\section{Discussion}
\label{sec:discussion}
In this work, we identified a bias against high spatial frequencies in convolutional GANs, and then proposed an approach for manipulating this bias and reducing its adverse effects. Being aware of this bias is of paramount importance when using GANs in applications concerned with intricate patterns, \eg de-nosing and reconstruction in medical or satellite imaging. Moreover, our findings suggest that the information carried by high frequencies is much more likely to be missed by GANs, a critical consideration when using GANs for data-augmentation or in semi-supervised learning. We also observed that the spatial frequency bias primarily affects GAN's generator and not its discriminator. This gives the discriminator an advantage which can be the root of certain instabilities in GAN training. Investigating this connection between the spatial frequency bias and unstable GAN training, as well as extending Theorem 1 to incorporate the effect of various normalizations and stabilization techniques, are interesting directions for future research. Finally, we would like to emphasize that while FSG exposes certain behaviors of the spatial frequency bias, and can be used to match the bias of GANs to the known biases of a dataset, it is not a proper solution to the bias, in that it does not remove the bias, rather exploits it. Devising proper solutions for the bias is yet another exciting and important direction for future research.

\bibliography{refs}
\bibliographystyle{iclr2021_conference}
\clearpage

\begin{appendices}
\section{Proof of Proposition 1}
\begin{proof}
Let $f(W): \Rb^n \rightarrow \Rb$ denote a scalar output of a Conv layer of a finite ReLU-CNN with an n-dimensional parameter space ($\mathcal{W}$), at a parameter vector $W$, and at a latent input $H^1 \in \Rb^{d_0\times d_0}$. We want to show that for all $W\in \mathcal{W}$, except a measure zero set, there exists a neighborhood of $W$ in which either $f(\hat{W}) \geq 0$ for all members $\hat{W}$ or $f(\hat{W}) \leq 0$ for all members. To do so, it suffices to show the following set has measure zero in $\mathcal{W}$:
\begin{align}
    G = \{ W\in{\mathcal{W}} \quad | \quad \forall \mathcal{N}(W) \quad \exists U,V \in \mathcal{N}(W): f(U) < 0 < f(V) \}
\end{align}
where $\mathcal{N}(W)$ denotes a neighborhood of $W$. Incidentally, since $f$ is a continuous function, this is the set of zero-crossings of $f(W)$ over the parameter space $\mathcal{W}$. Since $f(W)$ has a finite set of neurons, it has a finite set of ReLU activations, and therefore at any $W$ it will be equivalent to one of a finite set of possible polynomials on $\mathcal{W}$ (corresponding to the finitely many binary permutations of ReLUs). A polynomial function on $\Rb^n$ to $\Rb$ has a measure zero set of zero-crossings in the parameter space~\citep{caron2005zero}. Therefore, a finite set of such polynomials also has a measure zero set of zero-crossings, which concludes that $G$ has measure zero. Finally, note that the same argument holds for any scalar output $f$ of the CNN, at any spatial location in any layer, and given the finite number of these outputs, the measure of $G$ for all outputs $f$s is also zero, which completes the proof.
\end{proof}

\section{Proof of Theorem 1}
\begin{proof}
Let us consider a single 2D filter $F^l \in \Rb^{k_l \times k_l}$ in the $l$-th Conv layer. Since the spatial dimension of the layer's output is $d_l$, a filter $G^l \in \Rb^{d_l \times d_l}$ can sufficiently model any spectrum in the layer's output space. So we can write $F^l$ as a restriction of $G^l$, that is, the multiplication of $G^l$ with the 2D pulse $P \in \Rb^{d_l \times d_l}$ of area $k_l^2$:
\begin{align}
    &F^l = P.G^l\\
    &P(x, y) =
    \begin{cases}
    1 & 0\leq x,y < k_l \\ 0 & \textrm{otherwise}
    \end{cases}
\end{align}
Therefore, by convolution theorem, the spectrum of $F^l$ is equal to the spectrum of $G^l$ convolved by the spectrum of $P$:
\begin{align}
    \FT{F^l} = \FT{P.G^l} = \FT{P}*\FT{G^l} = \Sinc*\FT{G^l}
\end{align}
where $\FT{.}$ denotes the $d_l$-point DFT, $*$ is the circular convolution, and $\Sinc$ is the aliased sinc function:
\begin{align}
    \Sinc(u,v) = \frac{\sin(\frac{\pi uk_l}{d_l})\sin(\frac{\pi vk_l}{d_l})}{\sin(\frac{\pi u}{d_l})\sin(\frac{\pi v}{d_l})} e^{-j\pi(u+v)\frac{(k_l - 1)}{d_l}} \quad d_l, k_l \in \Nb \quad \textrm{and} \quad 1 < k_l \leq d_l
\end{align}
Now, let $U=\FT{F^l}(u_0, v_0)$ and $V=\FT{F^l}(u_1, v_1)$ be two spatial frequency components on the spectrum of the filter, located at the $(\frac{u_0}{d_l}, \frac{v_0}{d_l})$ and $(\frac{u_1}{d_l}, \frac{v_1}{d_l})$ frequencies respectively. We are interested in the correlation between these two components. Note that we are defining $G^l$ as the hypothetical unrestricted filter, such that any possible filter that can be learnt during training becomes a restriction of $G^l$. Therefore, without loss of generality, we can assume $G^l$ has independent spatial frequency components with variance $\sigma^2$. Considering these assumptions, we can find the complex correlation coefficient~\citep{park2018fundamentals} between $U$ and $V$ as follows:
\begin{align}
    &\Cov{U, V} = \Cov{\Sinc * \FT{G^l} (u_0, v_0), \Sinc *  \FT{G^l} (u_1, v_1) }\\
    &= \Cov{\sum_{u,v}\Sinc(u,v)\FT{G^l} (u_0-u, v_0-v), \sum_{\hat{u},\hat{v}}\Sinc(\hat{u},\hat{v}).\FT{G^l} (u_1-\hat{u}, v_1-\hat{v})}\\
    &= \sum_{u,v}\sum_{\hat{u},\hat{v}} \Sinc(u,v)\Sinc^*(\hat{u},\hat{v})\Cov{\FT{G^l} (u_0-u, v_0-v), \FT{G^l} (u_1-\hat{u}, v_1-\hat{v})}\\
    &= \sum_{u,v} \Sinc(u,v)\Sinc(u_0-u_1-u, v_0-v_1-v)\Var{\FT{G^l}(u_0-u, v_0-v)}\\
    &= \sigma^2 \Sinc * \Sinc (u_0-u_1, v_0-v_1) = \sigma^2 d_l^2 \Sinc (u_0-u_1, v_0-v_1)\\\\
    &\Var{U} = \Var{\Sinc * \FT{G^l} (u_0, v_0)}\\
    &= \Var{\sum_{u,v} \Sinc(u,v) \FT{G^l} (u_0-u, v_0-v)}\\
    &= \sum_{u,v} |\Sinc(u,v)|^2 \Var{\FT{G^l} (u_0-u, v_0-v)} = \sigma^2 \sum_{u,v} |\Sinc(u,v)|^2 = \sigma^2 d_l^2 k_l^2\\\\
    &\textrm{corr}(U, V) = \frac{\Cov{U, V}}{\sqrt{\Var{U}\Var{V}}} = \frac{\Sinc(u_0-u_1, v_0-v_1)}{k_l^2}
\end{align}
\end{proof}
Note that all the expectations are taken over the probability space of all unrestricted 2D filters of a specific output dimension $d_l$, that is the probability space of $G^l$ when considered as a random variable, and that $\Sinc^*$ represents the complex conjugate of $\Sinc$.

\clearpage

\section{GAN Generated Samples}
\begin{figure*}[h]
    \centering
    \includegraphics[trim=0 2345 0 1240, clip, width=\textwidth]{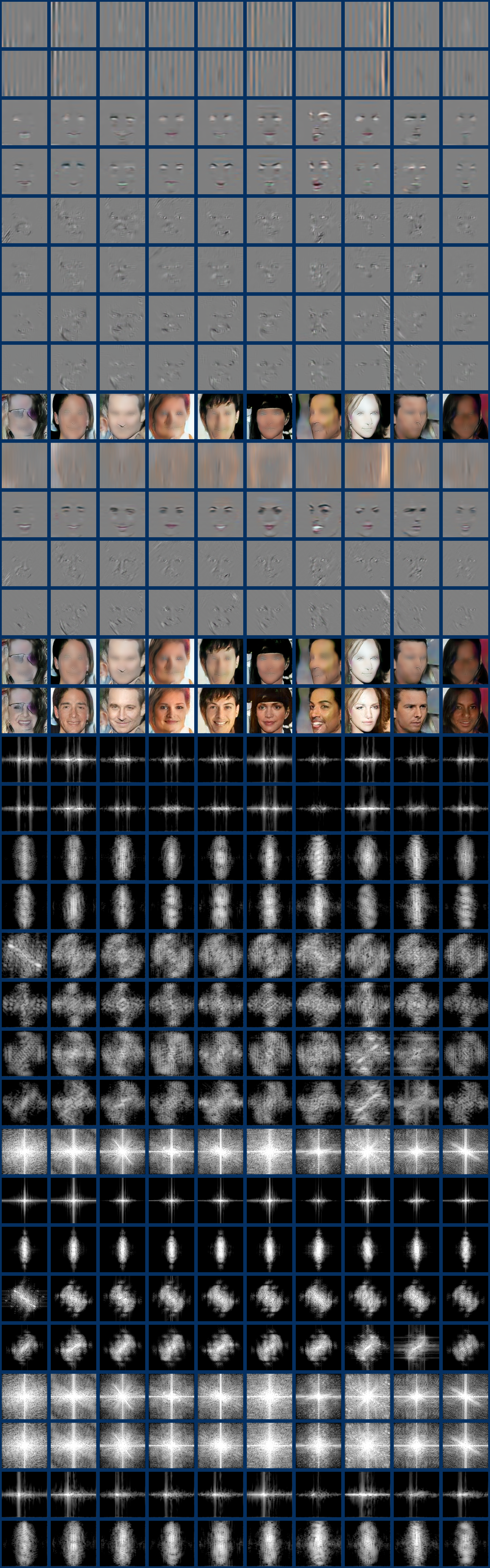}
    \caption{Each column corresponds to one sample of a WGAN-GP enhanced by adding multiple FSGs, trained on CelebA. The first four rows, from top to bottom, show the outputs of the FSGs with $(\hat{u}_t, \hat{v}_t)$ at $(\frac{1}{16}, 0)$, $(0, \frac{1}{16})$, $(-\frac{1}{16}, \frac{1}{16})$, and $(\frac{1}{16}, \frac{1}{16})$ respectively. The fifth row shows the main generator, and the final row the output of the compound generator (sum of all the preceding rows).}
    \label{fig:fsg16_wgan_samples}
\end{figure*}

\begin{figure*}[h]
    \centering
    \includegraphics[trim=0 0 0 0, clip, width=\textwidth]{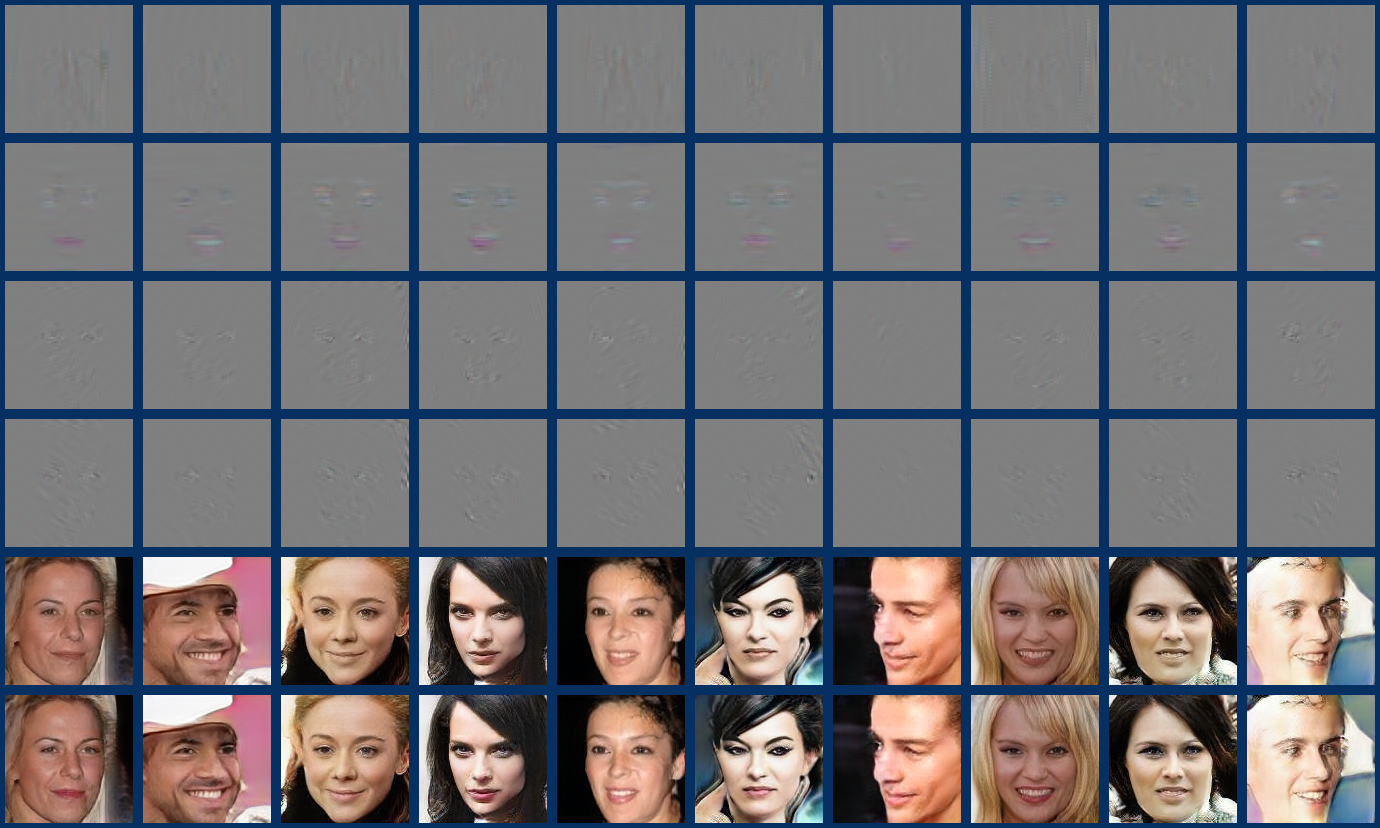}
    \caption{Each column corresponds to one sample of a PG-GAN enhanced by adding multiple FSGs, trained on CelebA. The first four rows, from top to bottom, show the outputs of the FSGs with $(\hat{u}_t, \hat{v}_t)$ at $(\frac{1}{16}, 0)$, $(0, \frac{1}{16})$, $(-\frac{1}{16}, \frac{1}{16})$, and $(\frac{1}{16}, \frac{1}{16})$ respectively. The fifth row shows the main generator, and the final row the output of the compound generator (sum of all the preceding rows).}
    \label{fig:fsg16_pggan_samples}
\end{figure*}

\begin{figure*}[h]
    \centering
    \includegraphics[trim=0 0 0 0, clip, width=\textwidth]{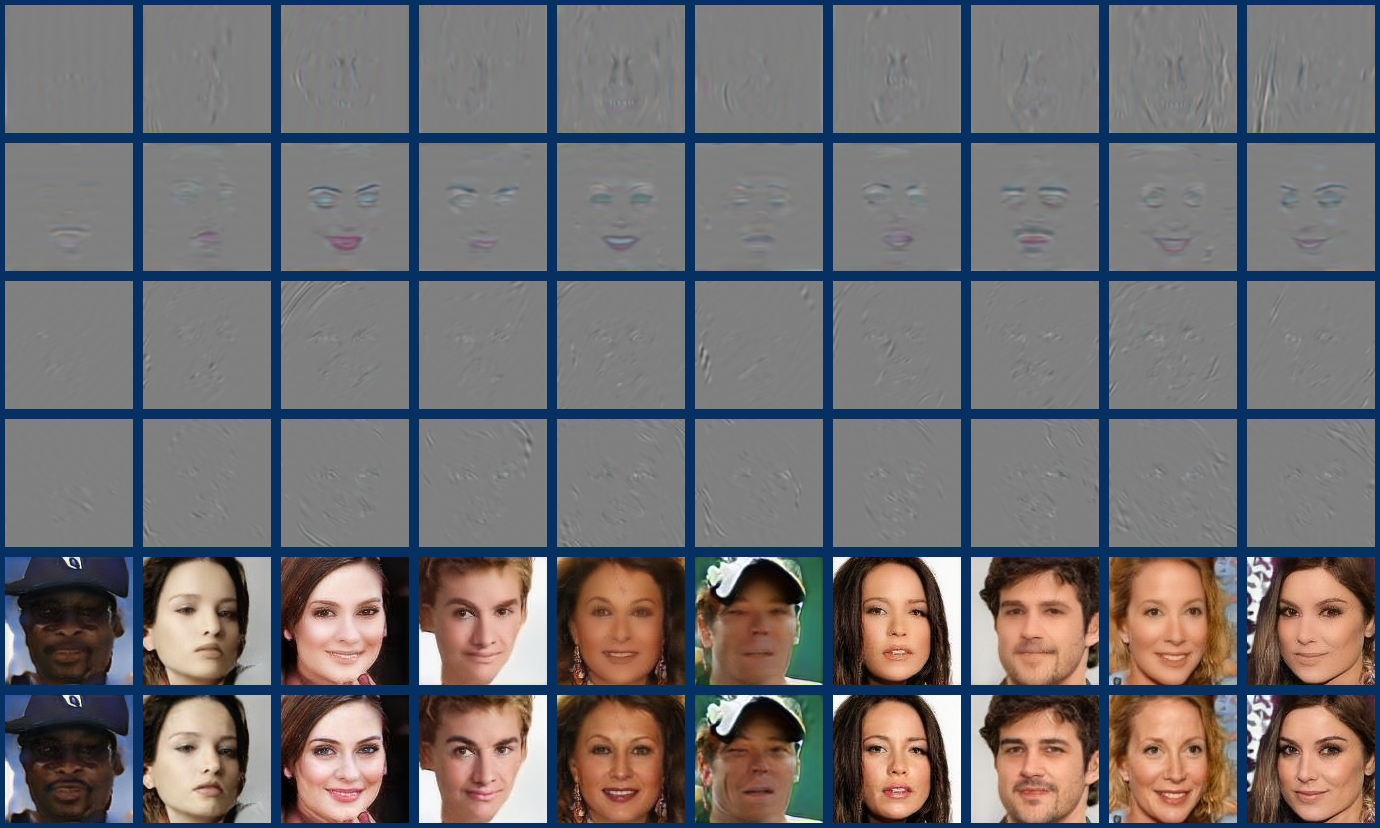}
    \caption{Each column corresponds to one sample of a StyleGAN2 enhanced by adding multiple FSGs, trained on CelebA. The first four rows, from top to bottom, show the outputs of the FSGs with $(\hat{u}_t, \hat{v}_t)$ at $(\frac{1}{16}, 0)$, $(0, \frac{1}{16})$, $(-\frac{1}{16}, \frac{1}{16})$, and $(\frac{1}{16}, \frac{1}{16})$ respectively. The fifth row shows the main generator, and the final row the output of the compound generator (sum of all the preceding rows).}
    \label{fig:fsg16_stylegan2_samples}
\end{figure*}

\begin{figure*}[t]
    \centering
    \subfloat[WGAN-GP on SCelebA]{
        \centering
        \includegraphics[trim=0 0 0 0, clip, width=0.48\textwidth]{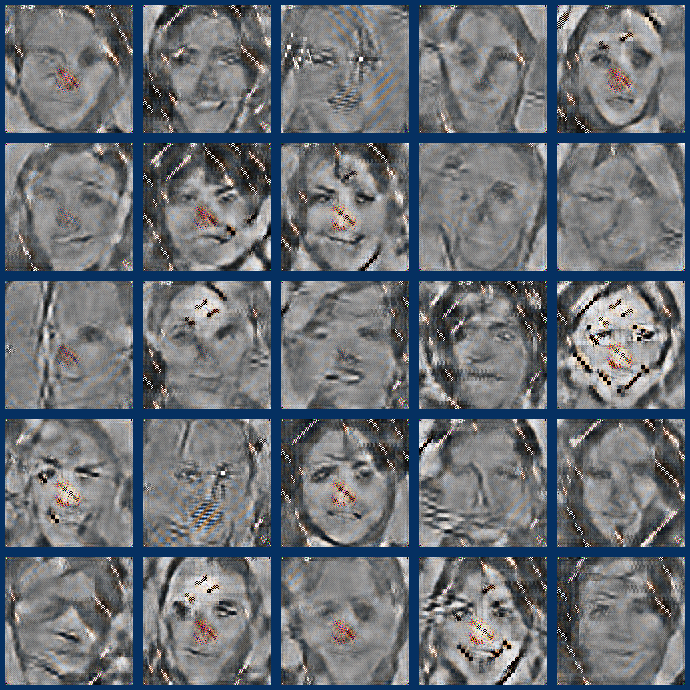}
        }
    \subfloat[WGAN-GP-FSG on SCelebA]{
        \centering
        \includegraphics[trim=0 0 0 0, clip, width=0.48\textwidth]{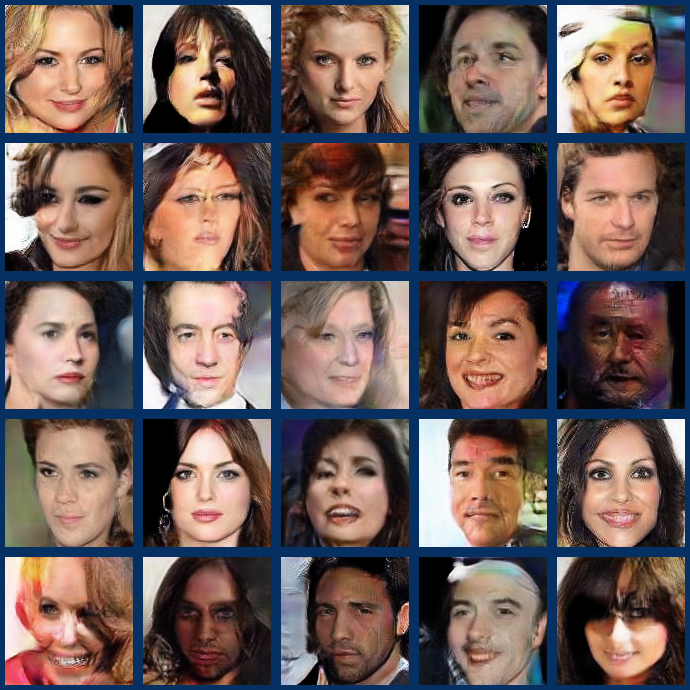}} \quad
    \subfloat[WGAN-GP on SBedrooms]{
        \centering
        \includegraphics[trim=0 0 0 0, clip, width=0.48\textwidth]{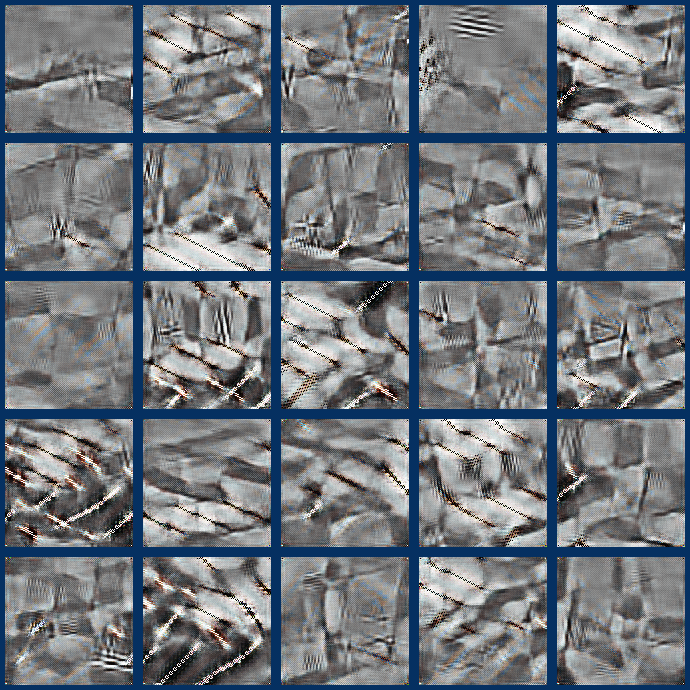}} \hspace{0.005in}
    \subfloat[WGAN-GP-FSG on SBedrooms]{
        \centering
        \includegraphics[trim=0 0 0 0, clip, width=0.48\textwidth]{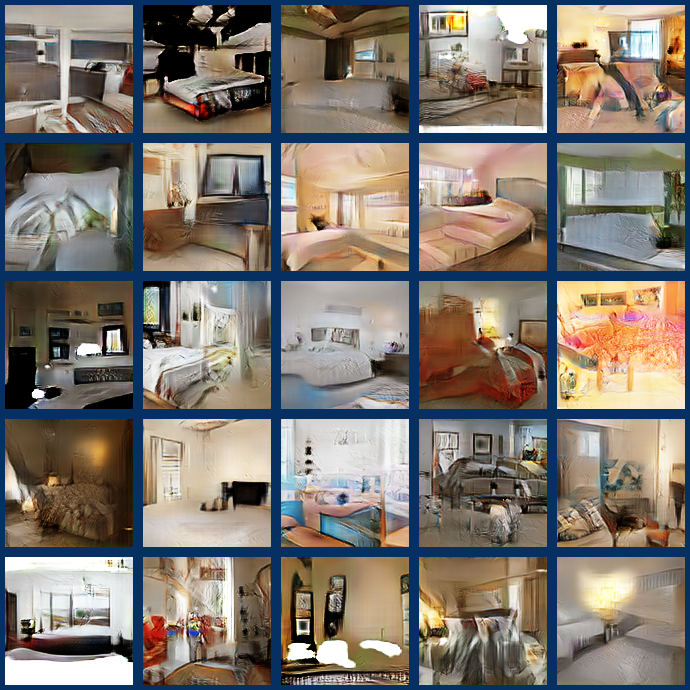}}
    \caption{WGAN-GP random samples compared to WGAN-GP-FSG on SCelebA and SBedrooms. Samples are re-shifted for visualization.}
    \label{fig:wgan_fsg_samples}
\end{figure*}

\begin{figure*}[t]
    \centering
    \subfloat[PG-GAN on SCelebA]{
        \centering
        \includegraphics[trim=0 0 0 0, clip, width=0.48\textwidth]{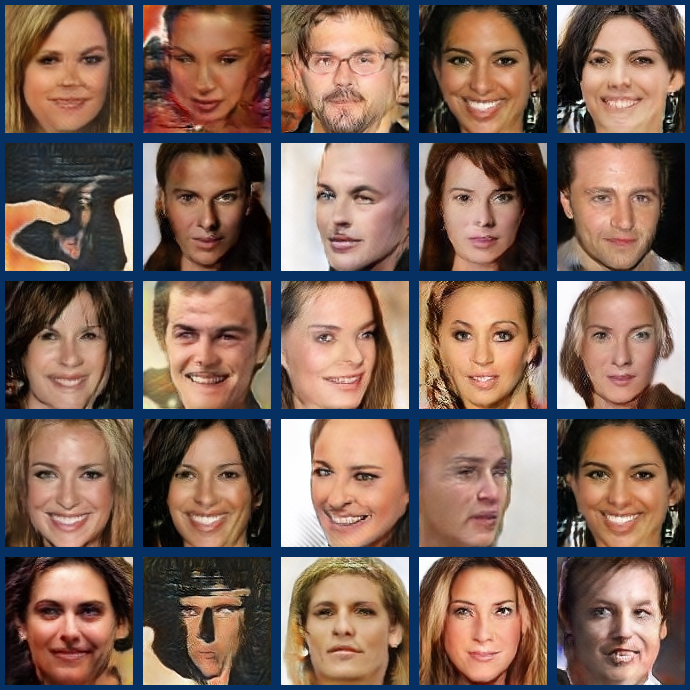}
        }
    \subfloat[PG-GAN-FSG on SCelebA]{
        \centering
        \includegraphics[trim=0 0 0 0, clip, width=0.48\textwidth]{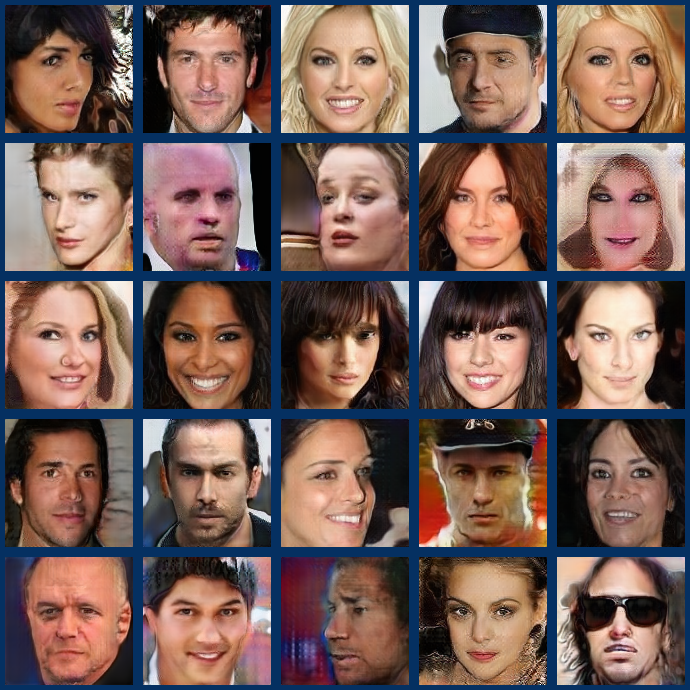}} \quad
    \subfloat[PG-GAN on SBedrooms]{
        \centering
        \includegraphics[trim=0 0 0 0, clip, width=0.48\textwidth]{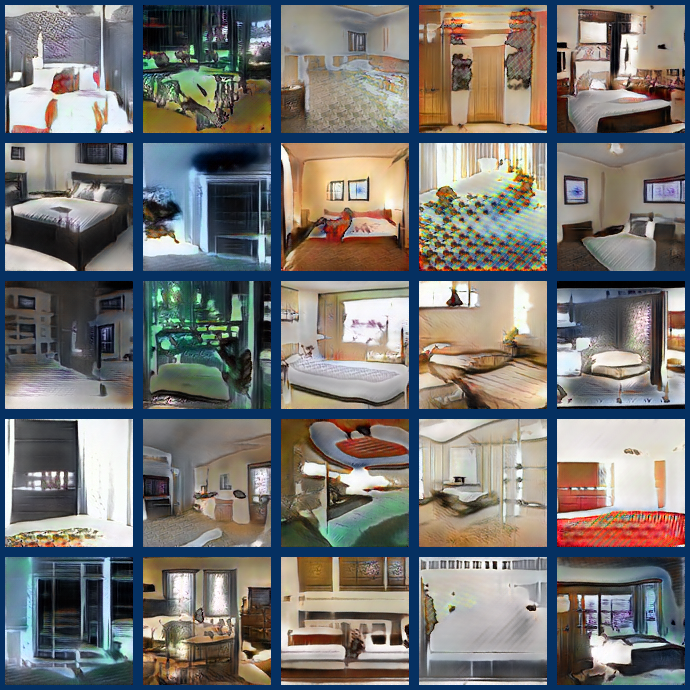}} \hspace{0.005in}
    \subfloat[PG-GAN-FSG on SBedrooms]{
        \centering
        \includegraphics[trim=0 0 0 0, clip, width=0.48\textwidth]{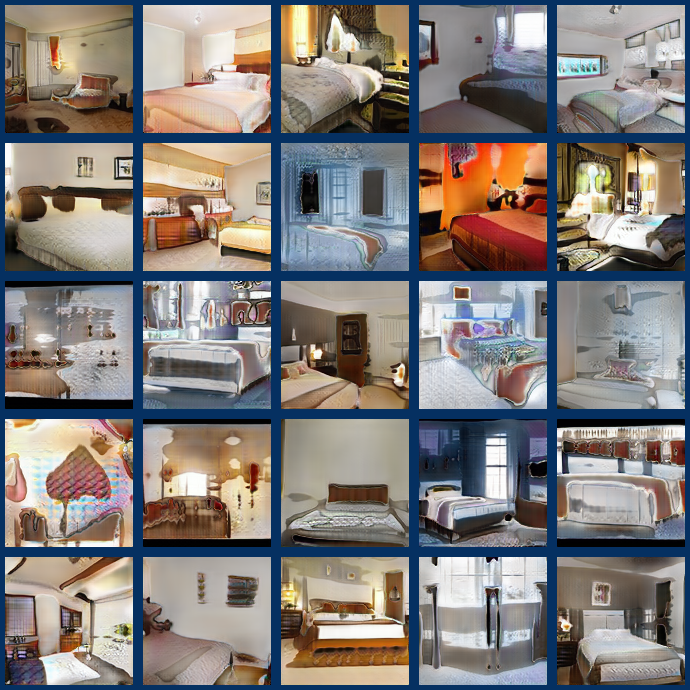}}
    \caption{PG-GAN random samples compared to PG-GAN-FSG on SCelebA and SBedrooms. Samples are re-shifted for visualization.}
    \label{fig:pggan_fsg_samples}
\end{figure*}

\begin{figure*}[t]
    \centering
    \subfloat[StyleGAN2 on SCelebA]{
        \centering
        \includegraphics[trim=0 0 0 0, clip, width=0.48\textwidth]{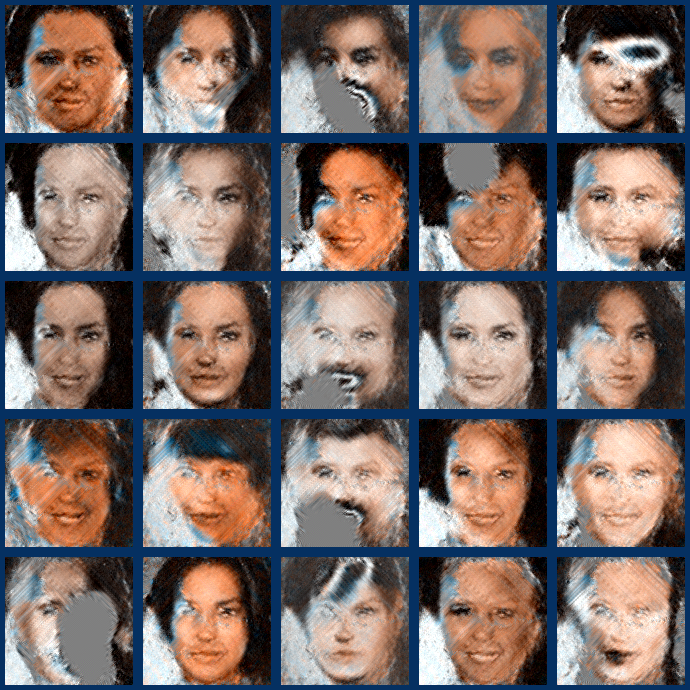}
        }
    \subfloat[StyleGAN2-FSG on SCelebA]{
        \centering
        \includegraphics[trim=0 0 0 0, clip, width=0.48\textwidth]{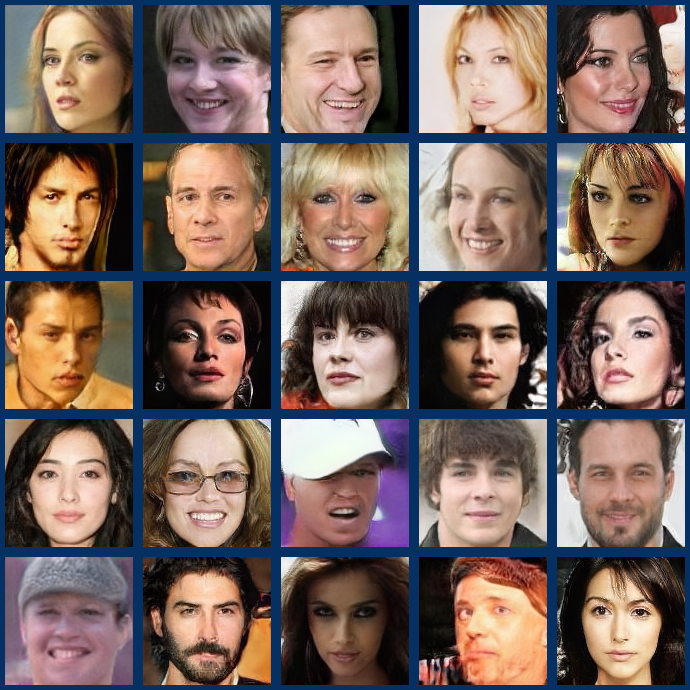}} \quad
    \subfloat[StyleGAN2 on SBedrooms]{
        \centering
        \includegraphics[trim=0 0 0 0, clip, width=0.48\textwidth]{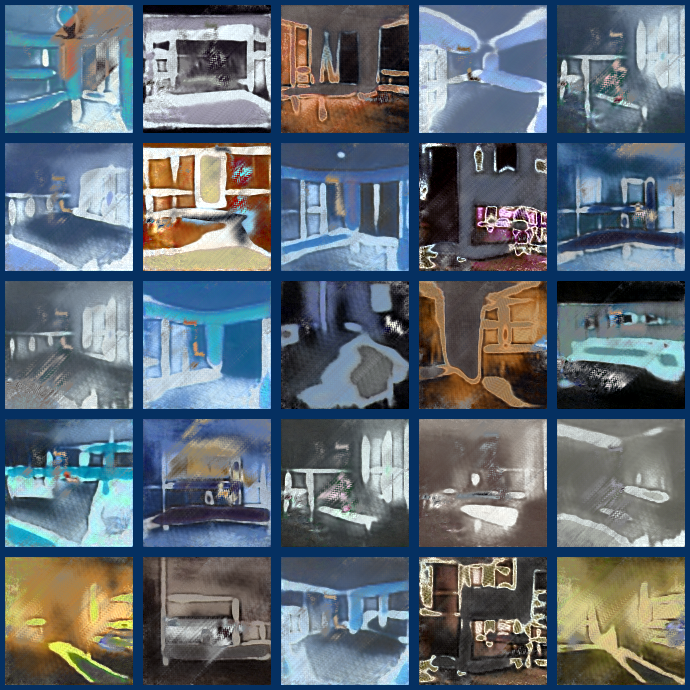}} \hspace{0.005in}
    \subfloat[StyleGAN2-FSG on SBedrooms]{
        \centering
        \includegraphics[trim=0 0 0 0, clip, width=0.48\textwidth]{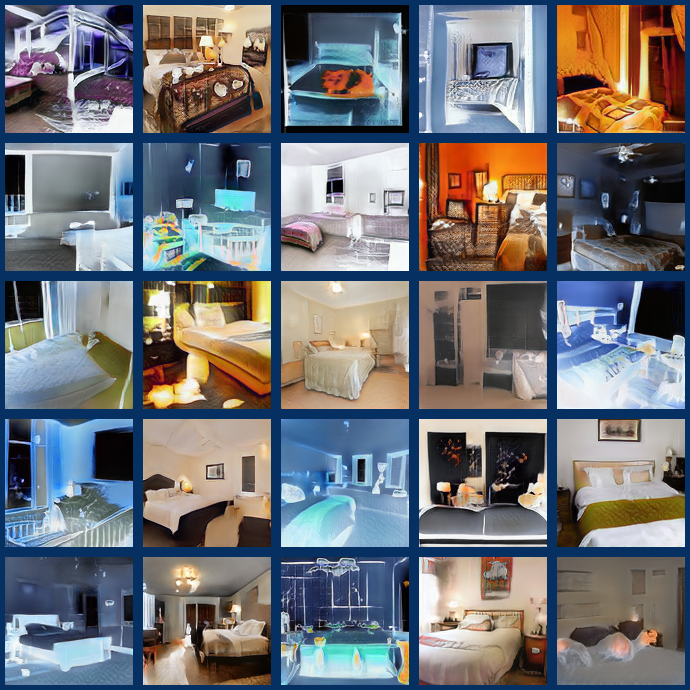}}
    \caption{StyleGAN2 random samples compared to StyleGAN2-FSG on SCelebA and SBedrooms. Samples are re-shifted for visualization.}
    \label{fig:stylegan2_fsg_samples}
\end{figure*}

\clearpage

\section{Experiment Details}
For WGAN-GP, we use the network in Table~\ref{tab:wganbn_net}, and train according to the specifications of~\citet{gulrajani2017improved} for 200 epochs. For PG-GAN, we use the network in Table~\ref{tab:pggan_net} and train according to the specifications of~\citet{karras2018progressive} for 10M images. For StyleGAN2, we use the network and training setup of config-e in~\citet{karras2020analyzing}, except that we use the same progression of CNN filter channel sizes as in Table~\ref{tab:pggan_net}, and train for 10M images. On the $1024\times1024$ fractal dataset of~\Fig{}~\ref{fig:koch}, the exact network and setup of config-e is used for the large-scale StyleGAN2.

In multiple FSGs experiments, for WGAN-GP, each FSG has the same network as WGAN-GP itself, except it has only up-sampling instead of the last layer, so that FSGs have less capacity than the main generator. For PG-GAN and StyleGAN2, each FSG shares all the Conv blocks of the corresponding main generator network except the two outer most blocks, which are replaced by a single Conv block followed by up-sampling to the output spatial dimension, that is, the FSGs are essentially branching out of the main generator at the start of the $64\times64$ Conv layer. In StyleGAN2, the added FSGs share the mixing input to the last block of the main generator.

For FID calculation, 50K images are sampled per distribution in CelebA and SCelebA experiments, and 10K images are sampled in LSUN-Bedroom and SBedrooms experiments, due to the limited number of available true images. For spectrum visualization, 1K images are sampled.

\setlength{\tabcolsep}{7pt}

\begin{table*}[h]
\begin{center}
\caption{WGAN-GP's networks. $G(z)$ and $D(x)$ refer to the generator and discriminator respectively, where $z\in {[-1, 1]}^{128}$ is the randomly sampled latent vector and $x\in [-1, 1]^{128\times128\times3}$ is the input image.}
\label{tab:wganbn_net}
\vspace{\baselineskip}
\begin{tabular}{lllll}
    \toprule
    Operation & Kernel & Strides & Output Shape & Activation \\
    \midrule
    \textbf{G(z)}: $z \sim \textrm{Uniform}(-1,1)$ & & & 128 &\\
    Fully Connected & & & $8\times8\times512$ & ReLU\\
    Nearest Upsample & & & $16\times16\times512$ &\\
    Convolution & $5\times5$ & $1\times1$ & $16\times16\times256$ & ReLU\\
    Nearest Upsample & & & $32\times32\times256$ & -\\
    Convolution & $5\times5$ & $1\times1$ & $32\times32\times128$ & ReLU\\
    Nearest Upsample & & & $64\times64\times128$ & -\\
    Convolution & $5\times5$ & $1\times1$ & $64\times64\times64$ & ReLU\\
    Nearest Upsample & & & $128\times128\times64$ & -\\
    Convolution & $5\times5$ & $1\times1$ & $128\times128\times3$ & Tanh\\
    \midrule
    \textbf{D(x)}: x & & & $128\times128\times3$ & \\
    Convolution & $5\times5$ & $2\times2$ & $64\times64\times64$ & Leaky ReLU\\
    Convolution & $5\times5$ & $2\times2$ & $32\times32\times128$ & Leaky ReLU\\
    Convolution & $5\times5$ & $2\times2$ & $16\times16\times256$ & Leaky ReLU\\
    Convolution & $5\times5$ & $2\times2$ & $8\times8\times512$ & Leaky ReLU\\
    Fully Connected & & & $1$ & \\
    \bottomrule
\end{tabular}
\end{center}
\end{table*}

\begin{table*}[t]
\begin{center}
\caption{PG-GAN's networks. $G(z)$ and $D(x)$ refer to the generator and discriminator respectively, where $z\in \Rb^{512}$ is the randomly sampled latent vector and $x\in [-1, 1]^{128\times128\times3}$ is the input image.}
\label{tab:pggan_net}
\vspace{\baselineskip}
\begin{tabular}{lllll}
    \toprule
    Operation & Kernel & Strides & Output Shape & Activation \\
    \midrule
    \textbf{G(z)}: $z \sim \Nc(0,1)$ & & & 512 &\\
    Convolution & $4\times4$ & $1\times1$ & $4\times4\times512$ & Leaky ReLU\\
    Convolution & $3\times3$ & $1\times1$ & $4\times4\times512$ & Leaky ReLU\\
    
    Nearest Upsample & & & $8\times8\times512$ &\\
    Convolution & $3\times3$ & $1\times1$ & $8\times8\times512$ & Leaky ReLU\\
    Convolution & $3\times3$ & $1\times1$ & $8\times8\times512$ & Leaky ReLU\\
    
    Nearest Upsample & & & $16\times16\times512$ &\\
    Convolution & $3\times3$ & $1\times1$ & $16\times16\times256$ & Leaky ReLU\\
    Convolution & $3\times3$ & $1\times1$ & $16\times16\times256$ & Leaky ReLU\\
    
    Nearest Upsample & & & $32\times32\times256$ &\\
    Convolution & $3\times3$ & $1\times1$ & $32\times32\times128$ & Leaky ReLU\\
    Convolution & $3\times3$ & $1\times1$ & $32\times32\times128$ & Leaky ReLU\\
    
    Nearest Upsample & & & $64\times64\times128$ &\\
    Convolution & $3\times3$ & $1\times1$ & $64\times64\times64$ & Leaky ReLU\\
    Convolution & $3\times3$ & $1\times1$ & $64\times64\times64$ & Leaky ReLU\\
    
    Nearest Upsample & & & $128\times128\times64$ &\\
    Convolution & $3\times3$ & $1\times1$ & $128\times128\times32$ & Leaky ReLU\\
    Convolution & $3\times3$ & $1\times1$ & $128\times128\times32$ & Leaky ReLU\\
    Convolution & $1\times1$ & $1\times1$ & $128\times128\times3$ &\\
    \midrule
    \textbf{D(x)}: x & & & $128\times128\times3$ & \\
    Convolution & $1\times1$ & $1\times1$ & $128\times128\times32$ & Leaky ReLU\\
    Convolution & $3\times3$ & $1\times1$ & $128\times128\times32$ & Leaky ReLU\\
    Convolution & $3\times3$ & $1\times1$ & $128\times128\times64$ & Leaky ReLU\\
    Box Downsample & & & $64\times64\times64$ &\\
    
    Convolution & $3\times3$ & $1\times1$ & $64\times64\times64$ & Leaky ReLU\\
    Convolution & $3\times3$ & $1\times1$ & $64\times64\times128$ & Leaky ReLU\\
    Box Downsample & & & $32\times32\times128$ &\\
    
    Convolution & $3\times3$ & $1\times1$ & $32\times32\times128$ & Leaky ReLU\\
    Convolution & $3\times3$ & $1\times1$ & $32\times32\times256$ & Leaky ReLU\\
    Box Downsample & & & $16\times16\times256$ &\\
    
    Convolution & $3\times3$ & $1\times1$ & $16\times16\times256$ & Leaky ReLU\\
    Convolution & $3\times3$ & $1\times1$ & $16\times16\times512$ & Leaky ReLU\\
    Box Downsample & & & $8\times8\times512$ &\\
    
    Convolution & $3\times3$ & $1\times1$ & $8\times8\times512$ & Leaky ReLU\\
    Convolution & $3\times3$ & $1\times1$ & $8\times8\times512$ & Leaky ReLU\\
    Box Downsample & & & $4\times4\times512$ &\\
    
    Minibatch stddev & & & $4\times4\times513$ &\\
    Convolution & $3\times3$ & $1\times1$ & $4\times4\times512$ & Leaky ReLU\\
    Convolution & $4\times4$ & $4\times4$ & $1\times1\times512$ & Leaky ReLU\\
    Fully Connected & & & $1$ & \\
    \bottomrule
\end{tabular}
\end{center}
\end{table*}
\end{appendices}

\end{document}